\documentclass[conference]{IEEEtran}
\IEEEoverridecommandlockouts

\pdfoutput=1
\usepackage{cite}
\usepackage{amsmath,amssymb,amsfonts}
\usepackage{algorithmic}
\usepackage{graphicx}
\usepackage{textcomp}
\usepackage{xcolor}
\usepackage{tcolorbox}

\usepackage{algorithm}
\usepackage{algorithmic}

\usepackage[hyphens]{url}  

\newcommand{\ie}{i.e.}
\newcommand{\eg}{e.g.}
\usepackage{amsmath}
\usepackage{amssymb}
\newcommand{\ignore}[1]{}
\newcommand*\Mname{\textsc{ColPer }}
\newcommand\zl[1]{{\color{magenta}{\textbf{\{zl: {\em#1}\}}}}}
\newcommand{\zztitle}[1]{\vspace{2pt}\noindent\textbf{#1.}}
\usepackage{amssymb}
\DeclareMathOperator*{\argmax}{arg\,max}
\DeclareMathOperator*{\argmin}{arg\,min}
\usepackage[algo2e, ruled,vlined]{algorithm2e}
\usepackage{multirow}
\usepackage{tabularx}
\usepackage{extarrows}
\newcommand{\jxu}[1]{{\color{orange}{\textbf{\{Jason: {\em#1}\}}}}}
\newcommand{\revised}[1]{{\color{black}{{#1}}}}
\newcommand{\shephred}[1]{{\color{black}{{#1}}}}
\usepackage{authblk}

\def\BibTeX{{\rm B\kern-.05em{\sc i\kern-.025em b}\kern-.08em
    T\kern-.1667em\lower.7ex\hbox{E}\kern-.125emX}}
\begin{document}

\title{On Adversarial Robustness of Point Cloud Semantic Segmentation}
\ignore{
\author{\IEEEauthorblockN{Jiacen Xu}
\IEEEauthorblockA{\textit{University of California, Irvine}\\
jiacenx@uci.edu}
\and
\IEEEauthorblockN{Zhe Zhou}
\IEEEauthorblockA{\textit{Fudan University}\\
zhouzhe@fudan.edu.cn}
\and
\IEEEauthorblockN{Boyuan Feng, Yufei Ding}
\IEEEauthorblockA{\textit{University of California, Santa Barbara}\\
\{boyuan, yufeiding\}@ucsb.edu}
\and
\IEEEauthorblockN{Zhou Li}
\IEEEauthorblockA{\textit{University of California, Irvine}\\
zhou.li@uci.edu}
}
}
\author[1]{Jiacen Xu}
\author[2]{Zhe Zhou}
\author[3]{Boyuan Feng}
\author[3]{Yufei Ding}
\author[1]{Zhou Li}

\affil[1]{\textit{University of California, Irvine}}
\affil[2]{\textit{Fudan University}}
\affil[3]{\textit{University of California, Santa Barbara}}

\maketitle
\pagestyle{plain}
\begin{abstract}
Recent research efforts on 3D point cloud semantic segmentation (PCSS) have achieved outstanding performance by adopting neural networks. However, the robustness of these complex models have not been systematically analyzed. Given that PCSS has been applied in many safety-critical applications like autonomous driving, it is important to fill this knowledge gap, especially, how these models are affected under adversarial samples. 
As such, we present a comparative study of PCSS robustness. First, we formally define the attacker's objective under performance degradation and object hiding. Then, we develop new attack by whether to bound the norm. We evaluate different attack options on two datasets and three PCSS models. We found all the models are vulnerable and attacking point color is more effective. With this study, we call the attention of the research community to develop new approaches to harden PCSS models.

\end{abstract}


\section{Introduction}
\label{sec:intro}



Accurate and robust perception are keys to the success of autonomous systems, with applications on autonomous driving, autonomous food delivery, etc. 
\revised{
The main equipments for perception include LiDAR (Light Detection and Ranging) sensor, which uses laser light to measure distances, camera, etc.~\cite{chen20203d}. 
These sensors can model the environment as dense, geo-referenced and accurate \textit{3D point cloud}, which is a collection of 3D points that represent the surface geometry.}

To process the point cloud data, deep-learning models based on Convolutional Neural Network (CNN) and Graph Convolutional Network (GCN) have been extensively leveraged~\cite{li2019deepgcns,landrieu:hal-01801186,Wei_2020_CVPR}.
Due to their usage in safety-critical applications, a number of works have attempted to generate adversarial examples against such deep-learning models, which migrate the existing attacks against 2D images, like Fast Gradient Sign Method (FGSM) \cite{goodfellow2014explaining}, iterative FGSM (iFGSM) \cite{kurakin2016adversarial}, Projected Gradient Descent (PGD) \cite{madry2017towards}, and Carlini \& Wagner (CW)~\cite{carlini2017towards}, to the 3D point cloud setting~\cite{yang2019adversarial,xiang2019adv,Zhou_2020_CVPR,liu2019extending,DBLP:journals/corr/abs-1904-00923, zheng2019pointcloud, xie2017adversarial, tsai2020robust, huang2022shape, liu2022boosting, jin2023pla}. However, we found all of them focused on the task of \textit{objection recognition}, which identifies objects within an image or a video stream and assigns \textit{one} class label to the \textit{whole} point cloud. 
\ignore{
Some papers~\cite{kim2020minimal} provides new attack methods such as attacking the classification ability of point cloud-based networks and claim that their methods can be used on semantic segmentation tasks while they just validate on the classification datasets such as ModelNet40~\cite{wu20153d} and ScanObjectNN~\cite{uy2019revisiting} datasets.
}
Though objection recognition is an important task, \textit{point cloud semantic segmentation (PCSS)} is probably more relevant to real-world autonomous systems, as it is the process of labeling \textit{each point} in a 3D point cloud with a semantic class label, such as ``ground'', ``building'', ``car'', or ``tree'' and aims to classify \textit{many} objects in a real-world \textit{scene}. 
The major use cases of PCSS include obstacle avoidance and boundary detection. As far as we know, the work from Zhu et al.~\cite{zhu2021adversarial} is the only one attacking PCSS models, but it is only tested under the setting of LiDAR sensing, and only one outdoor dataset is evaluated. Hence, the robustness of PCSS models under the adversarial samples has not been systematically explored, and there is an urgent need to answer questions like which PCSS model design is more robust and in what setting (e.g., indoor or outdoor) the attack is more likely to succeed, to guide the development of assured autonomy.

\ignore{
In addition, the previous adversarial attacks concentrate on outdoor setting which is commonly used in auto driving but the indoor setting which has plenty of applications like the electric wheelchair~\cite{mohamed2021indoor}, virtual reality~\cite{tiator2020using}, and augmented reality~\cite{zhang2020slimmer} is ignored. Our paper fills the missing brick in the field.}

\begin{figure}[ht]
    \centering
    \includegraphics[width=\columnwidth]{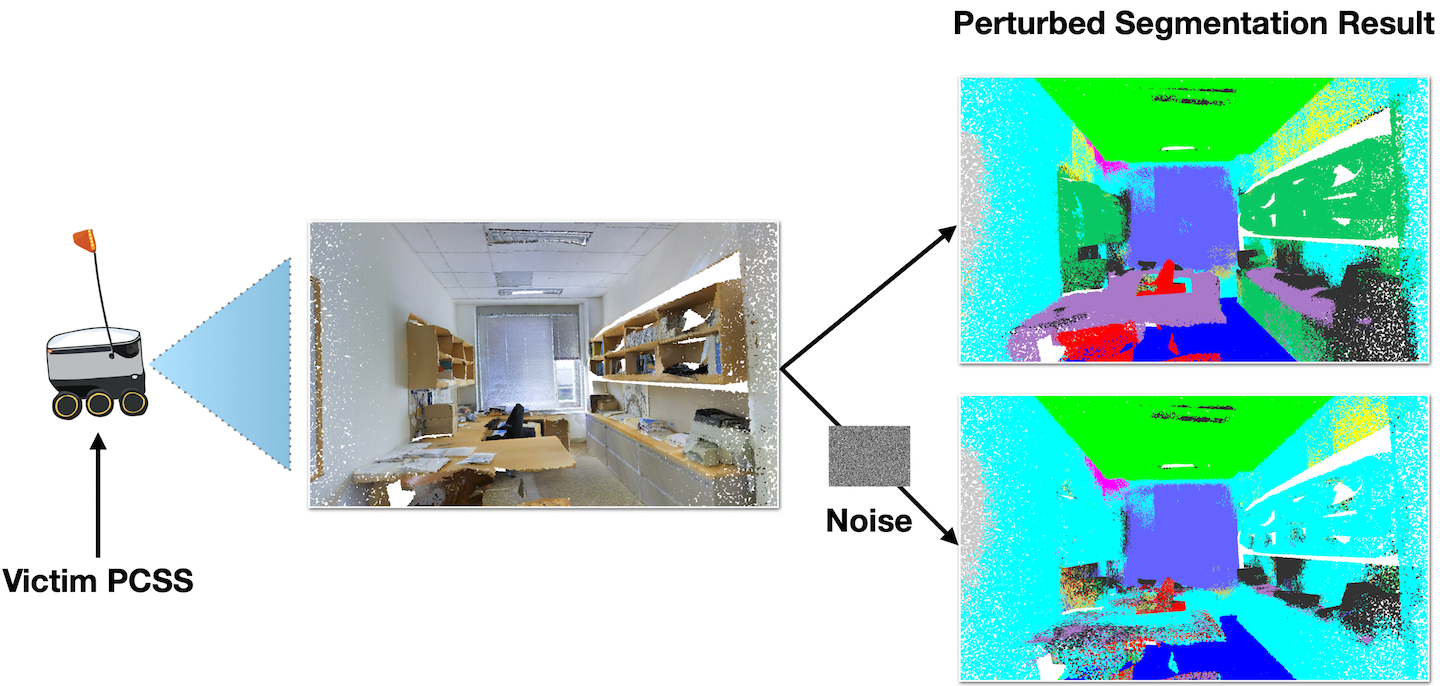}
    \caption{One example of color-based perturbation under the object hiding attack setting. Different objects are colored differently. Multiple objects (desk, chair, and bookcase) are misclassified as wall after the attack.
    }
    \label{fig:intro_example}
\end{figure}

Yet, answering these questions is non-trivial if directly applying the adversarial perturbation against 2D images or point cloud object recognition to PCSS. 
First, in the segmentation task, the class label is assigned to \textit{every} point, and the segmentation result of a point is also determined by its surrounding points, which increases the uncertainty of the attack outcome. Second, 
various data pre-processing procedures such as point sampling have been applied by the PCSS models, and the attack accuracy can be impacted by them. 
Finally, perturbation is only performed on the point coordinates by prior works, but there are other point features used for semantic segmentation. For instance, 9 features are included in a point cloud of the S3IDS dataset~\cite{xu2020grid} and 6 features are included in Semantic3D dataset~\cite{hackel2017semantic3d}. 
Whether and how the features undermine the robustness of the PCSS models have not been studied. 
  

\zztitle{Our Study}
We perform the first systematic and comparative analysis on the robustness of PCSS, by developing a \textit{holistic} attack framework that incorporate different attack configurations. We first convert attacker's objectives under \textit{object hiding attack} and \textit{performance degradation attack} into the forms that can be solved through optimization. 
Under each objective, we develop a norm-bounded attack method adjusted from \textit{PGD} attack~\cite{madry2017towards} and a norm-unbounded attack adjusted from \textit{CW} attack~\cite{carlini2017towards}, to compare their effectiveness. 
Previous attacks against point cloud \textit{all} focused on perturbing the point coordinates. However, we found their effectiveness is questionable under PCSS, as point sampling can make the attack outcome hard to control. 
As such, we exploit the point \textit{features} like color, and adjust the attack methods accordingly.
Hence, our attack framework supports \textit{8} attack configurations (object hiding/performance degradation $\times$ norm-bounded/norm-unbounded $\times$ coordinate-based/color-based), which enables a comprehensive analysis.




\zztitle{Evaluation Results}
We evaluate the attack framework against three popular PCSS models, including ResGCN-28~\cite{li2019deepgcns},  PointNet++~\cite{qi2017pointnet++} and RandLA-Net~\cite{hu2020randla}, as they represent different directions in processing point cloud. For the datasets, we use S3IDS and Semantic3D, representing both indoor and outdoor scenes. Below we highlight key findings:
\textbf{(1)} 
We compare the perturbation on point features (color in particular) against point coordinates, and our result shows that point features are more vulnerable. 
\textbf{(2)} Under the performance degradation attack, we found all tested PCSS models are vulnerable, with the norm-unbounded attack being more effective (e.g., dropping the segmentation accuracy \textit{from 85.90\% to 6.75\%} when attacking ResGCN-28). In the meantime, the perturbation added to the original point cloud is still small. Notably, PointNet++, ResGCN-28, and RandLA-Net belong to very different model families,
suggesting our developed attacks are universally effective.
\textbf{(3)} For the object hiding attack, as the attacker needs to select the source objects and determine what they should be changed to, the selection makes a big difference, as some objects are easier to manipulate (\eg, changing board to wall in S3IDS). Figure~\ref{fig:intro_example} shows an example of the attack. 
\textbf{(4)} The outdoor scenes are similarly vulnerable comparing to the indoor scenes (e.g., accuracy drops from 98.25\% to 16\% when RandLA-Net is used to segment Semantic3D point cloud).


Overall, our study demonstrates that the robustness of the deep-learning models under PCSS is questionable, and we outline a few directions for improving their robustness.

\zztitle{Contributions} 
1) We develop a holistic framework to enable various attack configurations against PCSS models and extend the previous attacks that are coordinate-based to color-based. 2) We evaluate different attack configurations against three types of PCSS models and indoor and outdoor datasets. Our code is released in a GitHub repository\footnote{PointSecGuard: \url{https://github.com/C0ldstudy/PointSecGuard}.}.


\ignore{
\zztitle{Roadmap}
Section~\ref{sec:background} discusses the background and the related works from deep learning on point clouds and adversarial examples. We formulate the problem of adversarial attacks against PCSS in Section~\ref{sec:problem}. The design of attack methods is elaborated in Section~\ref{sec:design}. We show the results in Section~\ref{sec:evaluation}. In Section~\ref{sec:discussion}, we discuss the potential defenses and limitations of this work, and Section~\ref{sec:conclusion} concludes the paper. 

}
\section{Background and Related Works}
\label{sec:background}


\subsection{Deep Learning on Point Cloud}
\label{subsec:dl}

3D point cloud generated by sensors has become a popular medium to represent the environment interacted by autonomous systems.
Two primary tasks have been explored with point clouds, namely objection recognition (or classification), and semantic segmentation. We focus on the second task.

To process the data stored in a point cloud, early works transformed the data into regular 3D voxel grids or images for the conventional CNN, which makes the data unnecessarily voluminous. PointNet~\cite{qi2017pointnet} addressed this issue by using a shared Multilayer Perceptron (MLP) on every individual point, and a global max-pooling to convert the input into a fixed-length feature vector. Since then, variations like PointNet++~\cite{qi2017pointnet++}, So-net~\cite{li2018so}, PointCNN~\cite{li2018pointcnn}, KPConv~\cite{thomas2019kpconv}, and PointNeXt~\cite{qian2022pointnext} have been proposed to use sophisticated modules and hierarchical architectures to aggregate local neighborhood information and extract local structure.
Besides CNN, DeepGCN~\cite{li2019deepgcns} shows that GCN can be leveraged to process point clouds. It solves the gradient vanishing problem when models become deeper, as the data are sparse in the geometry space but nodes in adjacency have strong relations.
To further reduce the overhead of handling large point clouds like the outdoor Semantic3D dataset, RandLA-Net~\cite{hu2020randla} leverages random point sampling and local feature aggregation, and shows 200× speedup. 
In this work, we evaluate the point cloud models from the aforementioned three directions, including PointNet++ under CNN,  ResGCN under GCN, and RandLA-Net under random point sampling. 

Recently, some techniques are proposed to improve the performance of semantic segmentation in particular, including contrastive boundary learning for better scene boundary analysis~\cite{tang2022contrastive} and multi-view aggregation to leverage the information from the associated 2D images~\cite{robert2022learning}. Other deep models like GAN~\cite{shu20193d} and transformer~\cite{mazur2020cloud, zhao2021point} have been used to process point cloud. In Section~\ref{sec:discussion}, we discuss the adaptability of our attacks to these new models.
%

\zztitle{Point Sampling}
The number of points included by a point cloud for semantic segmentation is often larger than object recognition. For example, 1,024 points are used to represent each object in the ModelNet40 dataset~\cite{wu20153d}, while 4,096 points and $10^8$ points are used to represent a scene in the S3DIS dataset~\cite{armeni20163d} and the Semantic3D dataset~\cite{hackel2017semantic3d}. Due to that the pre-processing and voxelization steps are computation-intensive, sampling the points in a point cloud becomes a standard approach by PCSS models, such as farthest point sampling~\cite{qi2017pointnet++}, and $k$-NN uniform sampling\cite{li2019deepgcns}, learning-based sampling~\cite{yang2019modeling,dovrat2019learning} and random sampling~\cite{hu2020randla}. We found point sampling makes it more difficult for the existing attacks that perturb point coordinates (to be discussed in Section~\ref{subsec:feature_coordinate}), which motivates us to explore the new attack methods that exploit point features.

\subsection{Adversarial Examples}
\label{subsec:examples}

The output of a point cloud model can be manipulated under adversarial examples. In the setting of 3D point cloud, existing works~\cite{yang2019adversarial,xiang2019adv,Zhou_2020_CVPR,DBLP:journals/corr/abs-1904-00923, zheng2019pointcloud, xie2017adversarial, tsai2020robust, huang2022shape, liu2022boosting, jin2023pla, kim2021minimal, liu2019extending}
took a gradient-based approach to generate adversarial examples.
\revised{
For example, Kim et al.~\cite{kim2021minimal} provide a unified framework to perturb and add points into a point cloud while minimizing the level of point manipulations.
Liu et al. ~\cite{liu2019extending} adapt the attack and defense methods against the 2D image to 3D point cloud.
}
GAN has also been used to create adversarial examples~\cite{zhou2020lg}.  However, these works aim to fool object recognition, which is a different task from this paper's focus.

Though attacks against semantic segmentation have been explored, most of the existing works~\cite{arnab2018robustness, nesti2022evaluating, nakka2020indirect} including FGSM \cite{goodfellow2014explaining}, iFGSM \cite{kurakin2016adversarial}, PGD \cite{madry2017towards}, and CW~\cite{carlini2017towards}, generate adversarial examples against \textit{2D images}, which have very different properties compared to 3D point clouds. 
The closest work comes from Zhu et al.~\cite{zhu2021adversarial}, which attacked LiDAR PCSS. Yet, only one outdoor dataset is evaluated and the perturbation is only applied to the coordinates. We believe the robustness of PCSS models has not been systematically evaluated, as PCSS can be used by applications other than autonomous driving (e.g., indoor navigation) and point features could also play an important role in addition to coordinates.
In fact, we found models like RandLA-Net and PointNet++ extensively leverage point features to boost the accuracy. 
In this paper, we make the \textit{first} attempt to comparatively analyze the robustness of PCSS, in order to fill this knowledge gap.




\subsection{Point Cloud Datasets}
\label{subsec:datasets}

To evaluate the performance of point cloud models, a number of public datasets have been released. For object classification, ModelNet~\cite{wu20153d}, ScanObjectNN~\cite{uy2019revisiting}, ShapeNet~\cite{chang2015shapenet} and PartNet~\cite{mo2019partnet} are widely used. For semantic segmentation, S3DIS~\cite{armeni20163d}, Semantic3D~\cite{hackel2017semantic3d} and KITTI~\cite{behley2019semantickitti} are the major datasets. Different datasets are created for object classification and semantic segmentation because the number of objects and labels differ: a scene for objection recognition has only one object and one label, while a scene for semantic segmentation usually has multiple objects and labels. It is also more challenging to perform semantic segmentation, as the classification result on one object can be impacted by the nearby objects in the same scene and some objects might only have a partial outline in the scene based on the separation of the point clouds.

In the paper, we select S3DIS \footnote{\revised{S3DIS is collected by Matterport scanners that are used for 3D space capture~\cite{armeni20163d}.}} and Semantic3D \footnote{\revised{Semantic3D is collected by high-resolution cameras and survey-grade laser scanners\cite{hackel2017semantic3d}.}} as the datasets to evaluate our attacks in both indoor and outdoor scenes. 
They both contain coordinate and color, enabling a fair assessment of the attack effectiveness on these two fields. 


\subsection{Threat Model}



\zztitle{Adversary's Goals} The attacker aims to change the perception results from the PCSS models deployed on autonomous systems like autonomous vehicles and delivery robots and the attack is also known as the evasion attack. 

In the real-world setting, for example, the attacker can realize two objectives by carefully introducing adversarial objects~\cite{zhu2021adversarial}, patches~\cite{tu2020physically}, or laser beam~\cite{jin2023pla} in the surrounding environment perceived by the sensors of the autonomous system. The attack consequences include rear-ending collision, sudden stop, abrupt driving direction change, etc.

We consider two attack scenarios. 
(1) \textit{Performance Degradation Attack}: this attack tampers the availability~\cite{papernot2018sok} of a PCSS model by forcing it to misclassify a large number of points, such that its prediction becomes entirely unreliable. 
(2) \textit{Object Hiding Attack}: this attack breaks the integrity~\cite{papernot2018sok} of a PCSS model by fooling it to classify points under an object as another object or the same as the background.

\zztitle{Adversary's Capabilities}
\revised{The adversary has white-box access to the victim PCSS model. In other words, the adversary has the read access to the model's structure and parameters and also access to the input of the autonomous system like the physical objects to be sensed. The attacker can generate a point cloud using the same PCSS model as the victim autonomous system and then perturb the points.

}
\ignore{
\zl{i dropped this statement, as ~\cite{zhu2021adversarial} showed it's possible to add points}
We do not consider adding or removing points because 
the number of points processed by PCSS is often a hyper-parameter that cannot be changed by the attacker. 
\zl{jason, is that true?}\jxu{I just checked. I think these models do not limit the point numbers. So the point number is not fixed. Maybe we can say: We ignore adding or removing nodes attack method because PCSS sampling technique filter the extra points which leads to the similar results of perturbing points. Besides, changing the number of point clouds are easy to be detected which are not useful in reality.}
}

\revised{
We assume the point coordinates or features (or both) can be perturbed by the attacker. We are motivated to investigate both fields because the recent point cloud datasets like S3IDS and Semantic3D contain both point coordinates and features like color. The authors collect data from both frequency-modulated continuous-wave (FMCW) LiDARs and high-resolution cameras, and apply \textit{multi-sensor fusion (MSF)} to assign features onto points. MSF is widely used by autonomous-driving companies like Google Waymo, Pony.ai, and Baidu Apollo to collect environmental information, and a recent work showed it is possible to generate adversarial object that is effective against both FMCW LiDAR and camera~\cite{cao2021invisible}, suggesting our adversarial samples are potentially realizable. 
Alternatively, high-end multi-spectral LiDAR can obtain both point coordinates and color, and our attack is expected to be effective as well.
}

\section{Problem Formulation}
\label{sec:problem}



In this section, we give a formal definition of point clouds and the attacker's goals. Table~\ref{tab:my_label} lists the main symbols used in this paper and their description.

\begin{table}[h]
    \caption{Main symbols used in the paper.}
    \centering
    \begin{tabular}{c|l}
        \textbf{Symbol} & \textbf{Description} \\ 
        \hline
        \hline
        $X$ & a point cloud \\
        $Y$ & the labels of all points in the point cloud \\
        $x_i$ & a point \\
        $y_i$ & a class label on $x_i$ \\
        $p_i$ & the coordinates of $x_i$ \\
        $c_i$ & the features of $x_i$ \\        
        $R$ & the perturbation values on the point cloud  \\
        $r_i$ & the perturbation values on $x_i$  \\
        $T$ & the set of point indices \\
        $\mathcal{F}_{\theta}(\cdot)$ & the model for PCSS\\
        $Z(\cdot)_i$ & the logits of the model's prediction \\
        $\mathcal{T}(\cdot)$ & the function selecting a subset of $X$ or $Y$\\
        $\mathcal{D}(\cdot)$ & the distance function \\
    \end{tabular}
    \label{tab:my_label}
\end{table}


A point cloud can be defined as a set of $N$ points, \ie, $\{p_i\}_{i=1}^N$, where each point $p_i = (pos_x, pos_y, pos_z)$ represents the 3D coordinates of a point. This basic form is usually sufficient for single-object recognition~\cite{qi2017pointnet,qi2017pointnet++}. 
We denote the features associated with a point $p_i$ as $c_i$, so a point cloud $X = \{x_i | i=1\dots N, x_i = \{p_i,c_i\} \}$ where $ c_i = (feat_1, feat_2, ..., feat_k)$ for $k$ features.

Below we formalize the two attack goals. 
First, we assume $\mathcal{F}_\theta: \mathcal{X} \xrightarrow{} \mathcal{Y}$ is the segmentation model which maps an input point cloud  
$X=\{x_i| i=1\dots N, x_i\in \mathcal{X} \}$
to the labels of \textit{all points} 
$Y=\{y_i| i=1\dots N, y_i\in \mathcal{Y} \}$. $\mathcal{X}$ is the universe of points, and $\mathcal{Y}$ is the universe of class labels, \eg, desk, wall, and chair. We design methods under both norm-bounded and norm-unbounded principles for the object hiding attack and the performance degradation attack.



\zztitle{Object Hiding Attack}
In this setting, the adversary chooses a subset of points $X_T=\{ x_i | i \in T, x_i \in X\}$, where $T$ is the set of indices, and perturbs $X_T$ to change their predicted labels to $Y_T=\{y_i| i \in T, y_i\in \mathcal{Y} \}$.
For a point $x_i=\{p_i, c_i\}$, we assume the attacker either perturbs its coordinates $p_i$ or (and) its features $c_i$. We treat coordinates and features separately as the perturbation methods have to be designed differently under their unique constraints. 
The perturbation values on the original point cloud $X$ can be represented as $R = \{r_i | i \in T\}$, and the new point cloud will be  $X' = \{x_i | i \notin T, x_i \in X\} + \{x_i + r_i | i \in T, x_i \in X\}$. 
Under coordinate-based perturbation, $r_i=\{r_{p_i}, 0^k\}$, where $r_{p_i}$ denotes the changes on the 3D coordinates. Under feature-based perturbation, $r_i=\{0^3, r_{c_i} \}$, where $r_{c_i}$ denotes the changes on the $k$ features. 

We first consider the norm-bounded attack, by which the attacker tries to minimize the difference between the predicted labels on $X_T$ and the targeted labels $Y_T$, while the perturbation is bounded by $\epsilon$. Hence, the attack goal can be formalized as:
\ignore{
\begin{equation}\label{eq:bounded_targeted}
    \argmin_{R} \mathcal{L}_\mathcal{F}(\mathcal{T}(\mathcal{F}_\theta (X'), T), Y_T), \;\; \text{s.t.} \;\;   \mathcal{D}(R) \leq \epsilon
\end{equation}
}
\begin{equation}\label{eq:bounded_targeted}
    \argmin_{R} \mathcal{L}_T(X', Y_{T}), \;\; \text{s.t.} \;\;   \mathcal{D}(R) \leq \epsilon
\end{equation}
where $\mathcal{D}(\cdot)$ is the distance function measuring the magnitude of the perturbation $R$ and $\mathcal{L}_\mathcal{T}(\cdot)$ is the adversarial loss that measures the effectiveness of the attack.
Notably, the attacker's goal is quite different from the attacks against single-object recognition~\cite{xiang2019adv}, where one label is assigned to the whole point cloud (\ie, the cardinality of $Y'$ is 1) and the number of points after perturbation can differ (\ie, $X'$ and $X$ have different cardinalities).

Under the norm-unbounded attack, the attacker tries to find the minimum perturbation values that can change the labels of $X_T$ to $Y_T$. Hence, the attacker's goal can be formalized as:
\begin{equation}\label{eq:unbounded_targeted_org}
    \argmin_{R} \mathcal{D}(R), \;\; \text{s.t.} \;\; \mathcal{T}(\mathcal{F}_\theta (X'), T) = Y_T
\end{equation}
where $\mathcal{T}(\cdot)$ selects the prediction results indexed by $T$ from $X'$.

Directly solving Equation~\ref{eq:unbounded_targeted_org} is difficult because the constraint $\mathcal{T}(\mathcal{F}_\theta (X'), T) = Y_T$ is non-differentiable~\cite{carlini2017towards}. 
As a result, we reformulate Equation \ref{eq:unbounded_targeted_org} to enable gradient-based optimization by introducing  $\mathcal{L}_T(\cdot)$ to replace this constraint, as shown in Equation \ref{eq:unbounded_targeted}. Besides, we add a smoothness penalty $\mathcal{S}(\cdot)$ to encourage the optimizer to keep $X'$ smooth, \ie, that the differences between the neighboring points are not drastic. 
\begin{equation}\label{eq:unbounded_targeted}
        \argmin_{R} \{ \mathcal{D}(R) + \lambda_1 \cdot \mathcal{L}_T(X', Y_{T}) + \lambda_2 \cdot \mathcal{S}(X') \}
\end{equation}
where
$\lambda_1$  and $\lambda_2$ are pre-defined hyper-parameters.

\begin{figure*}[h]
    \centering
    \includegraphics[width=\textwidth]{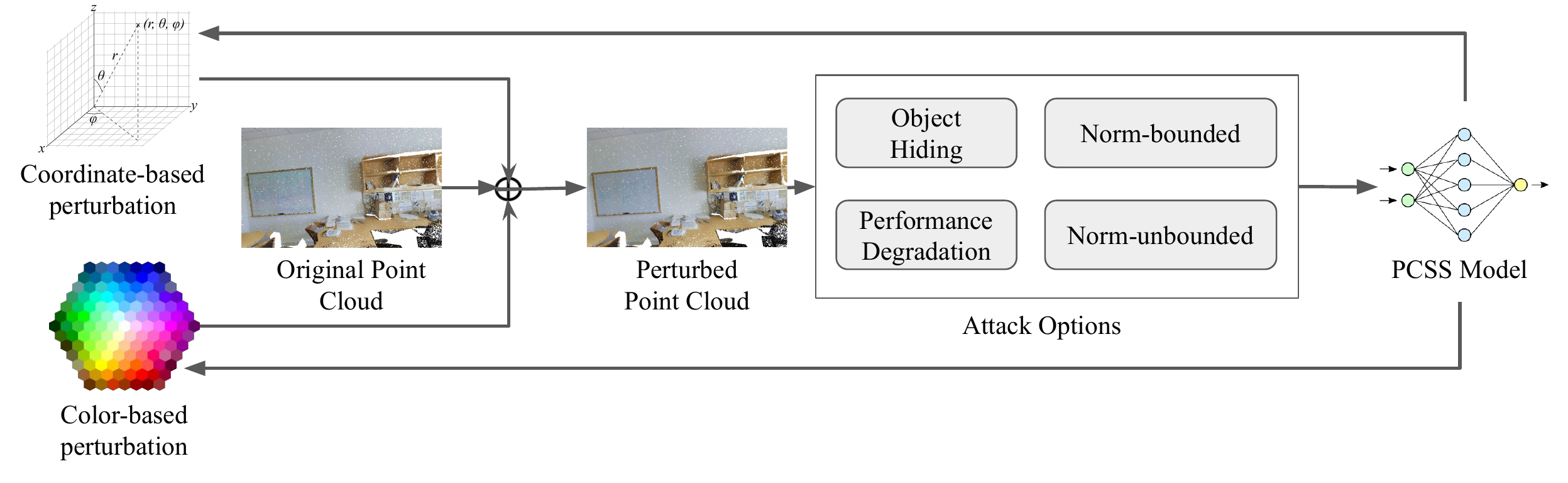}
    \caption{The attack workflow. 
    }
    \label{fig:structure}
\end{figure*}

\zztitle{Performance Degradation Attack}
In this setting, the adversary does not have a specific target $Y_T$, but just manipulates the prediction $\mathcal{F}_\theta (X')$ to be different from the ground-truth labels of all points in $X_T$ (termed $Y_{GT}$). Under norm-bounded attack, the attacker's goal can be adjusted from Equation \ref{eq:bounded_targeted} to:
\ignore{
\begin{equation}\label{eq:bounded_non_targeted}
    \argmax_{R} \mathcal{L}_\mathcal{F}(\mathcal{T}(\mathcal{F}_\theta (X'), T), Y_{GT}), \;\; \text{s.t.} \;\;   \mathcal{D}(R) \leq \epsilon
\end{equation}
}
\begin{equation}\label{eq:bounded_non_targeted}
    \argmax_{R}  \mathcal{L}_{NT}( X', Y_{GT}), \;\; \text{s.t.} \;\;   \mathcal{D}(R) \leq \epsilon
\end{equation}
where $\mathcal{L}_{NT}(\cdot)$ is the adversarial loss regarding $Y_{GT}$.

Under norm-unbounded attack, the attacker's goal can be adjusted from Equation \ref{eq:unbounded_targeted} to:
\begin{equation}\label{eq:unbounded_non_targeted}
        \argmin_{R} \{ \mathcal{D}(R) - \lambda_1 \cdot \mathcal{L}_{NT}( X', Y_{GT}) + \lambda_2 \cdot \mathcal{S}(X') \}
\end{equation}


\section{Attack Design} 
\label{sec:design}




As reviewed in Section~\ref{sec:background}, none of the prior attacks against PCSS consider the point features, so we highlight the design of  feature-based attacks here, which is supposed to be more resilient against point cloud sampling (see Section~\ref{subsec:dl}). In particular, we select the \textit{color} features as the perturbation target,  which turns $ c_i$ to $(color_r, color_g, color_b)$ for the three color channels, where $color_*$ is the pixel value and $*$ represents red, green, and blue.
\ignore{
We focus on the color channels as previous color-based attacks against 2D images show they are physically realizable, e.g., with carefully-printed stickers on the surface~\cite{eykholt2018robust}, and robust against surrounding illuminations, viewing angle, and distance~\cite{eykholt2018robust, zhao2019seeing}. On the other hand, perturbing the point coordinates at pixel granularity in the physical world could be more challenging~\cite{cao2019adversarial,tu2020physically}.
}



\subsection{Attack Components} 
\label{subsec:components}

Here we elaborate the distance function $\mathcal{D}(\cdot)$, adversarial loss functions $\mathcal{L}_{T}(\cdot)$ and $\mathcal{L}_{NT}(\cdot)$, smoothness penalty $\mathcal{S}(\cdot)$,  which are all listed in Section~\ref{sec:problem}.





\ignore{
\zl{it doesn't provide much more info}
\begin{figure*}[t]
    \centering
    \includegraphics[width=\linewidth]{pics/structure1.png}
    \caption{Overview of \Mname. The upper point clouds are the model input. The lower ones are the output with labels.\zl{re-draw it}}
    \label{fig:structure of pcgattack}
\end{figure*}
}

\zztitle{Distance Function}
When the color-based attack is chosen, we use $L_2$ distance to measure the magnitude of the perturbation, as shown by Equation~\ref{eq:color_dist}, because $L_2$ distance is commonly used by 2D image models~\cite{carlini2017towards}. Hence, the distance will be:
\begin{equation}
    \mathcal{D}(R) = \sum_{i \in T} ||r_{c_i}||_2^2 \label{eq:color_dist}
\end{equation}

As pointed out by Nicholas et al.~\cite{carlini2017towards}, optimization on $c_i$ encounters a box constraint, which is hard to solve. Hence, we map $c_i\in[a, b]$ to a new variable $W_i$, and perform optimization over $w_i$, as shown in Equation~\ref{eq:cw_tanh}.  
\begin{equation}\label{eq:cw_tanh}
    r_{c_i}=a+\frac{b-a}{2}(\text{tanh}(w_i)+1)
\end{equation}
where $\text{tanh}(\cdot)$ makes the gradient smoother and always falls in $[-1, 1]$, so the optimizer could find the right perturbation sooner. $a$ and $b$ are adjusted based on the normalized value range of the PCSS model.

When the coordinate-based attack is chosen, we use $L_0$ distance, i.e., how many points are changed, and the distance can be represented as:
\begin{equation}
    \mathcal{D}(R) = \sum_{i \in T} ||r_{p_i}||_0 \label{eq:coordinate_dist}
\end{equation}

We use $L_0$ distance  since other distance metrics like $L_2$ and $L_\infty$ yield different value ranges based on the input range of coordinates. Specifically, a color channel has a fixed range from 0 to 255 regardless of the point clouds, but a coordinate can have different value ranges for different point clouds. 
Equation~\ref{eq:cw_tanh} is also applied on the coordinate for the ease of optimization.


\ignore{
We use the same mapping function as Equation \ref{eq:cw_tanh} by just changing $\triangledown c_i$ to $\triangledown p_i$.
We set $a=-1$ and $b=1$ to adjust the perturbation range to $[-1, 1]$. \zl{we don't use the mapping for coordinates, jason, plz change the text}
}

\ignore{Notably, ~\cite{xiang2019adv} uses a more complex Chamfer distance to measure the point-wise distance. The Chamfer distance is not selected here because no point is removed or added and the computation on $L_2$ distance is fast. }

\zztitle{Smoothness Penalty}
The penalty is designed to make $X'$ smooth. In Equation \ref{eq:Smooth}, the distance between each point and its top $\alpha$ nearest neighbors is encouraged to be minimized.
\begin{align}\label{eq:Smooth}
    \mathcal{S}(X',\alpha) = \sum_{x'_i \in X'} \sum_{x'_j \in \text{Nei}(x'_i,\alpha)}(||x'_i-x'_j||_2)
\end{align}
where $\text{Nei}(x'_i,\alpha)$ returns the top $\alpha$ nearest neighbors. Noticeably, we consider all points rather than only points in $X_T$.

\zztitle{Adversarial Loss}
We use the logits (the output of the layer before the last softmax layer) of $\mathcal{F}_\theta$ to represent the adversarial loss for the object hiding attack. This loss encourages the optimizer to minimize the logits of the class rather than the target label.
\begin{equation}\label{eq:target_loss}
    \mathcal{L}_T(X', Y_T) = \sum_{\substack{ x'_i \in X' \\ y_i \in Y_T }}\max(\max\limits_{j\neq y_i}(Z(x'_i)_j)-Z(x'_i)_{y_i}, 0)
\end{equation}
where $Z(\cdot)_j$ represents the $j^{th}$ element of the logits of the adversarial example, and $Z(x'_i)_{y_i}$ means the target label's logits for $x'_i$. The largest logit that is not related to the target label is derived by $\max\limits_{j\neq y_i}(Z(x'_i)_j)$. 

\ignore{
\begin{equation}
    \mathcal{L}_t(X, X',Y_t) = Smooth(X, X', \alpha) + f_t(X', Y_t)
\end{equation}
\begin{equation}
    \mathcal{L}_{nt}(X, X',Y_{gt}) = Smooth(X, X', \alpha) + f_{nt}(X', Y_{gt})
\end{equation}
}

For performance degradation attacks, the adversarial loss is adapted to encourage the prediction of points to be any class other than the ground-truth classes. 
The loss function can be changed from Equation~\ref{eq:target_loss} to:
\begin{equation}\label{eq:nontarget_loss}
    \mathcal{L}_{NT}(X', Y_{GT}) = \sum_{\substack{ x'_i \in X' \\ y_i \in Y_{gt} }}
    \max(Z(x'_i)_{y_i}-\max\limits_{j\neq y_i}(Z(x'_i)_j), 0)
\end{equation}



\ignore{

\xjc{It seems that we do not need $\sum$}

In the non-target attack, the target label $t$ is set as the ground truth and we use the negative of the $Z(X'_t)_t - max(Z(X'_t)_j:j\neq t)$ for minimizing.
\begin{equation}
    f_{nt}(X', Y_{gt}) = max(Z(X')_{Y_{gt}} - max(Z(X')_j:j\neq t), 0)
\end{equation}

}




\begin{algorithm2e}[t]
\SetAlgoLined
\KwIn{the original point cloud $X$, the ground-truth labels $Y_{GT}$, the maximal number of iterations $Steps$, the target labels $Y_{T}$, the attack type $type=\{color, coordinate\}$, the attack boundary $\epsilon$, the target points mask $\mathcal{T}(\cdot)$ (all points for non-target attack), the targeted points $T$, the step size $\gamma$}
\KwOut{the adversarial example $X'$}
 $X_0 \leftarrow X, r \leftarrow X(type), i \leftarrow 1 $\;
 \While{$i \leq  Steps$}{
  $X^i_T(type)$ = $\mathcal{T}(r)$\;
  $\mathcal{T}(r)$ = $X^i_T(type) - X^{i-1}_T(type)$\;
  $X_T^i$ = $X_T^{i-1} + \mathcal{T}(r)$\;
  \uIf{object hiding attack}{
  $X_T^i = \text{clip}_{(-\epsilon, \epsilon)}(X_T^{i-1} \;
  - \gamma \cdot \text{sign}(\triangledown_{X_T}\mathcal{L}_T(X_T^{i-1}, Y_{T})))$\;
  }\uElseIf{performance degradation attack}{
  $X_T^i = \text{clip}_{(-\epsilon, \epsilon)}(X_T^{i-1} + \gamma \cdot \text{sign}(\triangledown_{X_T}\mathcal{L}_{NT}(X_T^{i-1}, Y_{GT})))$\;
  }

  \uIf{ $Converge$  }{
    return $X_T^i$\;
  }
  $i\leftarrow i+1$\;
 }
 return $X_T^i$\;
\caption{Norm-bounded attack.
}
\label{alg:algorithm_pgd}
\end{algorithm2e}

\subsection{Attack Workflow}
\label{subsec:workflow}



Our norm-bounded attack is adjusted from \textit{Projected Gradient Descent (PGD)}~\cite{madry2017towards} to PCSS\ignore{, and we use \PGD\ to denote it}. In essence, in each iteration, the attack adds noise to the perturbed point cloud of the previous iteration $X_T^{i-1}$ to derive $X_T^{i}$, following the goals defined in Equation~\ref{eq:bounded_targeted} (object hiding) and Equation~\ref{eq:bounded_non_targeted} (performance degradation). Then it clips the changes to $(-\epsilon, \epsilon)$ on $X_T^{i}$. Random initialization is used to set up $X_T^0$. To avoid the imbalance during the update, a sign operator is applied on the gradient. We set an upper bound of iterations as $\text{Steps}$. In each step, we use $\text{Converge}(\cdot)$ to determine if the adversarial example is satisfactory, based on the attacker's evaluation metrics, e.g., the dropped accuracy. 
\ignore{
For the object hiding attack, each step can be represented as:
\begin{equation}\label{eq:pgd_targeted_step}
   X_T^t = \text{clip}_{(-\epsilon, \epsilon)}(X_T^{t-1} - \gamma \cdot \text{sign}(\triangledown_{X_T}\mathcal{L}_T(X_T^{t-1}, Y_{T})))
\end{equation}
Where $t$ denotes the number of steps and a random initialization is used for $X_T^0$. $\text{clip}(\cdot)$ is the clip function to bound the perturbation to  $(-\epsilon, \epsilon)$. $\text{sign}(\cdot)$ is the sign operator. $\gamma$ is the step size. $\triangledown_{X_T}$ takes a gradient step in the direction of loss. 
For the performance degradation attack, each step can be adjusted from Equation~\ref{eq:pgd_targeted_step} to:
\begin{equation}\label{eq:pgd_non_targeted_step}
   X_T^t = \text{clip}_{(-\epsilon, \epsilon)}(X_T^{t-1} + \gamma \cdot \text{sign}(\triangledown_{X_T}\mathcal{L}_{NT}(X_T^{t-1}, Y_{GT})))
\end{equation}
}
Algorithm~\ref{alg:algorithm_pgd} lists the main steps. 


Our norm-unbounded attack is adjusted from \textit{Carlini and Wagner (CW)} attack~\cite{carlini2017towards}\ignore{, and we use \CW\ to denote it}. There are two main differences between our implementation with the norm-bounded attack.
First, each time when the target color is mapped to a new variable $W_i$ from Equation~\ref{eq:cw_tanh} ($a=0$ and $b=1$ for color), it utilizes the inverse function of Equation~\ref{eq:cw_tanh} before perturbing the original data points.
Second, instead of using attack boundary $\epsilon$~\cite{carlini2017towards}, we add a distance function item in the loss function under the goals defined in Equation~\ref{eq:unbounded_targeted} (object hiding) and Equation~\ref{eq:unbounded_non_targeted} (performance degradation) to optimize both the perturbation norm and attack success rate.
Like norm-bounded attack, we also bound the attack iterations by $\text{Steps}$. In each step, the gain over the attack is examined by $\text{Converge}(\cdot)$ as well. In addition for the norm-unbounded attack, if the gain does not increase in 10 steps, the perturbation will add random noise following the uniform distribution in $(0,1)$. When the adversarial example is invalid,  e.g., $c_i \notin [0,1]^3$, a new noise will replace the previous one. 


When the attacker chooses to perturb the coordinates, we select a set of the most impactful points and only perturb them, such that the $L_0$ criteria can be met. Especially, in each iteration $i$, we assume the points allowed to be perturbed is $X_i \subseteq X_T$.
After perturbation, the $n$ least impactful points are selected by the function $\text{MinImp}(X_i, n)$, and the point cloud is restored. The next iteration $i+1$, the perturbation will be only executed on $X_{i+1} = X_i \setminus \text{MinImp}(X_i, n)$. When the remaining points that can be perturbed are less than 10\% after a number of iterations, the point cloud will be perturbed without restoration. The equation below shows how the $n$ points are selected.
\begin{equation}\label{eq:minimp}
    \text{MinImp}(X_T^i, n)=\argmin_n g_n\cdot r_n
\end{equation}
where $g_n$ is the gradient and $r_n$ means the perturbation value. 

Our default setting either perturbs color or coordinate, but it can be readily adjusted to perturb both fields.
Specifically, our method generates the gradients of color and coordinate concurrently in each iteration during optimization, after they are pre-processed separately. 
The distance function of Equation~\ref{eq:coordinate_dist} and adversarial loss from Equation~\ref{eq:target_loss} and Equation~\ref{eq:nontarget_loss} are reused.
An alternate approach is to perturb them in turns at different iterations. However, we found this approach has a worse result, because the gradient updates of color and coordinate offset each other.




Noticeably, though our attacks are adjusted from PGD and CW, notable changes exist. For instance, our norm-bounded attack does not use the original cross-entropy loss directly and our norm-unbounded attack adds a new smoothness penalty. Besides, we extend the attacks by customizing data preprocessing procedures by Equation~\ref{eq:cw_tanh}.


In Section~\ref{subsec:settings}, the hyperparameter values used to evaluate both attacks are described.

\section{Evaluation}
\label{sec:evaluation}


\ignore{
In the paper, we plan to answer the following research questions (RQs):
\begin{itemize}

\item[\bfseries RQ2:] How robust the semantic segmentation models can be to keep the performance?

\item[\bfseries RQ3:] How hard to lead these models to predict one object to another?

\end{itemize}
}

In this section, we report our evaluation results on various attack settings, target models, and datasets. Each experiment is conducted to answer one or few research questions and the findings are highlighted in the end.

\subsection{Experiment Settings}
\label{subsec:settings}

\zztitle{Target Models}
We use the pre-trained PointNet++~\cite{qi2017pointnet++}, ResGCN-28~\cite{li2019deepgcns}, and RandLA-Net \cite{hu2020randla} as the target models to evaluate our attacks, mainly because their codes and pre-trained models are publicly available\footnote{PointNet++: \url{https://github.com/yanx27/Pointnet_Pointnet2_pytorch}}\footnote{ResGCN-28: \url{https://github.com/lightaime/deep_gcns_torch}}\footnote{RandLA-Net: \url{https://github.com/QingyongHu/RandLA-Net}}, and they represent different directions in the point cloud domain. In Section~\ref{sec:background}, we give an overview about their designs. Below, we elaborate on their details.

PointNet++ consists of 4 abstraction layers and 4 feature propagation layers with 1 voter number for its multi-angle voting.
The overall point accuracy and average Intersection-over-Union (aIoU) of the pre-trained PointNet++ on the indoor dataset S3DIS are 82.65\% and 53.5\% respectively, as reported in its GitHub repo.
ResGCN-28 uses dynamically dilated \emph{k}-NN and residual graph connections, and the pre-trained model configures \emph{k} to 16. 
It has 64 filters and 28 blocks with 0.3 drop-out rate and 0.2 stochastic epsilon for GCN. 
the accuracy and aIoU of the pre-trained ResGCN-28 model on S3DIS are 85.9\% and 60.00\%, as reported in its GitHub repo.
The pre-trained RandLA-Net downsamples large point clouds.
Its accuracy and mIoU are 88.00\% and 70.00\% on S3DIS, 94.8\% and 77.4\% on Semantic3D. 

    


    
    

\zztitle{Dataset}
We evaluate the attacks on two large-scale 3D datasets: an indoor dataset S3DIS~\cite{armeni2017joint} and an outdoor dataset Semantic3D~\cite{hackel2017semantic3d}. They have been extensively used as the benchmark for PCSS. The S3DIS dataset is composed of 3D point clouds with color channels, which were collected at 6 areas in 3 different buildings with 13 class labels. 
Each point cloud contains 4,096 points, and each point has 9 features. 
The point cloud data could be pre-processed differently by their segmentation models.
As for PointNet++, each point cloud is segmented into several parts, and the coordinate and color are normalized to $[0, 3]$ and $[0, 1]$ by themselves. As for ResGCN-28, the coordinate is normalized to $[-1, 1]$ while the color feature is normalized to $[0, 1]$ by itself.
RandLA-Net regenerates the point clouds with 40,960 points by randomly duplicating and selecting the points. The color feature is also normalized to $[0, 1]$  by itself.

We evaluate against the three models with the point clouds of Area 5, of which 198,220 point clouds (78,649,818 points) are included~\cite{armeni2017joint}.
\ignore{
\zl{this is confusing}
Notably, due to the pre-processing\footnote{\url{https://shapenet.cs.stanford.edu/media/indoor3d\_sem\_seg\_hdf5\_data.zip}} of the PointNet++, the total number of the point clouds of Area 5 is not the same. 
For a fair comparison, we use the whole point clouds of Area 5 for the two models.
}
For the performance degradation attack on S3DIS, we choose the point clouds with high accuracy (over 30\%) as targeted point clouds.
For the object hiding attack, we randomly selected the 100 point clouds in Office 33 of Area 5 when evaluating against PointNet++ and ResGCN-28. 
As RandLA-Net has different requirements of the point number, for each class (\eg, table), we randomly selected 100 point clouds in Area 5 which at least contain 500 points in the class.

The Semantic3D dataset contains 30 point clouds including coordinates, color channels, and intensity in 8 classes. Each point cloud has over $10^8$ points to compose a $160\times240\times30$ $m^3$ area. PointNet++ and ResGCN-28 are not designed to handle such big point clouds, so we only evaluate against RandLA-Net.

\zztitle{Evaluation Metrics}
We use the drop of accuracy and \textit{average Intersection over Union} (aIoU), \revised{which measures the overall accuracy across all classes}, as the indicators of the attack's effectiveness.
On a point cloud, the accuracy and aIoU are defined as follows: assuming the number of all points and correctly classified points are $N$ and $TP$, accuracy equals to $\frac{TP}{N}$. For a class $i$, the aIoU is defined as $\frac{TP_i}{FN_i+FP_i+TP_i}$, where $FN_i$, $FP_i$, $TP_i$ are the number of false negatives, false positives and true positives for the class-related points. Below, we will primarily show the accuracy and aIoU averaged over point clouds. 

For the object hiding attack, the drop in accuracy and aIoU only measure whether the classification results are changed, but they neglect whether the predictions are misled toward the target classes. Therefore, we define another metric, \textit{point success rate (PSR)}, as the ratio of \textit{points} that are correctly perturbed over all the attacked points in $X_T$. 
\ignore{When an adversarial sample needs to follow certain constraints to be considered as successful, we use another metric, \textit{sample success rate (SSR)}, to measure the ratio of the perturbed \textit{samples} that meet the criteria. 
}
Besides measuring the success rate of attacks, we are also interested in whether the segmentation results of points outside of $X_T$ are changed, so we compute the accuracy and aIoU on the subset of these points separately and call the metrics \textit{``out-of-band'' (OOB)} accuracy and aIoU. From the attacker's perspective, the drop of OOB accuracy and aIoU should be as low as possible.


\zztitle{Attack Hyper-parameters} 
We set $\text{Steps}$ to 50 and 1,000 for norm-bounded and norm-unbounded attacks. Both $\lambda_1$ and $\lambda_2$ used by the adversarial loss are set to 1 and 0.1 based on empirical analysis. The step size ($\gamma$) of norm-bounded attack is 0.01 while the Adam optimizer of norm-unbounded attack with $0.01$ learning rate ($lr$) is used. The nearest neighbor $\alpha$ for Equation~\ref{eq:Smooth} is set to 10. 
The batch sizes are set to 8 when attacking PointNet++ while to 1 when attacking ResGCN-28 and RandLA-Net.
For the performance degradation attack, we examine whether the accuracy is dropped below 7.6\% (\textit{i.e.,} 1/13, 13 classes) for S3IDS and 12.5\% (\textit{i.e.,} 1/8) for Semantic3D, which means the model's prediction is the same as random guessing. 
When the coordinate is attacked, $n$ least impactful points should be selected in each iteration, and we set $n$ to 100 during the experiment.  

 
\zztitle{Experiment Platform}
We run the experiments on a workstation that has an AMD Ryzen Threadripper 3970X 32-Core Processor and 256 GB CPU memory with 2 NVIDIA GeForce RTX 3090. Our attacks run on PyTorch 1.7.1 for the pre-trained PointNet++ and ResGCN-28, and Tensorflow 1.15 for the pre-trained RandLA-Net.

\ignore{
\zztitle{Code}
As described in Section~\ref{evaluation} describes, we use the pre-trained PointNet++~\cite{qi2017pointnet++}, ResGCN-28 under the DeepGCN family~\cite{li2019deepgcns}, and RandLA-Net \cite{hu2020randla} as the target models to evaluate \Mname, mainly because their codes and pre-trained models are publicly available
We will also release the code of our attacks on GitHub.
}

\subsection{Evaluation on the Attacked Fields}
\label{subsec:feature_coordinate}



\begin{table}[ht]
    \caption{The results of performance degradation attack on ResGCN-28. ``Acc'', ``Avg'' and ``Coord'' are short for ``Accuracy'', ``Average'' and ``Coordinate''.}
    \centering
    \begin{tabular*}{0.5\textwidth}{p{0.03\textwidth}p{0.03\textwidth}|p{0.05\textwidth}p{0.04\textwidth}p{0.04\textwidth}|p{0.05\textwidth}p{0.04\textwidth}p{0.04\textwidth}}
    \hline 

    \multirow{2}{*}{}& &  \multicolumn{3}{c}{Norm-unbounded}  &  \multicolumn{3}{|c}{Norm-bounded}\\     

    \textbf{Case} & & {$\mathbf{L_0}$} & \textbf{Acc} (\%) & \textbf{aIoU} (\%) & $\mathbf{L_0}$ & \textbf{Acc} (\%) & \textbf{aIoU} (\%) \\
    \hline
    \hline
    \multirow{3}{*}{Color} & Best & 496.00 & 0.24 & 0.12 & 496.00 & 0.07 & 0.04  \\
    & Avg & \textbf{1635.17} & 9.04 & 4.81 & \textbf{1130.04} & 12.13 & 6.71 \\
    & Worst & 4096.00 & 27.86 & 16.18 & 4096.00 & 67.65 & 51.12 \\
    \hline

    \multirow{3}{*}{Coord} & Best & 2396.00 & 8.74 & 4.57 & 496.00 & 1.61 & 0.81  \\
    & Avg & 4065.21 & 27.63 & 16.48 & 2993.71 & 16.49 & 9.21 \\
    & Worst & 4096.00 & 63.43 & 46.44 & 4096.00 & 47.88 & 31.47 \\
    \hline
    
    \multirow{3}{*}{Both} & Best & 496.00  & 3.54 & 1.80 & 496.00 & 8.15 & 4.25 \\
    & Avg & 2519.00 & 12.57 & 7.05 & 2407.00 & 31.14 & 21.85 \\
    & Worst & 4096.00 & 82.08 & 69.61 & 4096.00 & 94.26 & 89.15 \\
    \hline
    \end{tabular*}

    \label{tab:l0_attack}
\end{table}

As described in Section~\ref{subsec:examples}, the prior attacks against point cloud objection recognition and semantic segmentation all perturbed the coordinate field, leaving other channels like color feature unexplored. Hence, we first assess how the attacked fields impact the attack's effectiveness. The experiment is conducted on the S3IDS dataset under performance degradation attack and we show the results when ResGCN-28 is the target model. A similar trend is observed on other models.
 
\ignore{
In the different models, the range of coordinates is various based on how they separate the point clouds into small pieces which makes the $L_2$ norm attack unfair to compare. For example, in PointNet++ the coordinates in some point clouds range from 2.0 to 1.0 while others from 12.0 to 0.0.
To solve the problem, we follow the $L_0$ norm attack setting from CW attack~\cite{carlini2017towards} to compare how many fewer points are needed to mislead the target models in the performance degradation attack. 
We use PointNet++ and ResGCN-28 because RandLA-Net processes the point position features which makes the comparison not fair.  
}

Due to that the range of coordinates varies by PCSS models, $L_2$ distance is unsuitable to measure perturbation, as explained in Section~\ref{sec:design}. Hence, we use $L_0$ distance for the coordinate-based attack. For a fair comparison, we also change the $L_2$ distance used by the color-based attack to $L_0$. 
\ignore{
\zl{seems to be already described in attack hyper-params}
We set the ratio of the adversarial samples that perturb less than 10\% points (explained in Section~\ref{subsec:workflow}) of a point cloud and drop the segmentation accuracy below 7.69\% (random guessing) as the target in this experiment.
}

In Table~\ref{tab:l0_attack}, we show the best-case, average-case, and worst-case (``best'' and ``worst'' show the examples most vulnerable and robust against the attack) among the attacked point clouds in terms of accuracy, aIoU and $L_0$ distance (higher accuracy and aIoU are worse for attack). The results show that the average $L_0$ distance is significantly lower when perturbing color than perturbing coordinate and both (1130.04 for norm-bounded and 1635.17 for norm-unbounded under color, while more than 2000 in any other case). The accuracy and aIoU also observed a deeper drop for color-based perturbation. 

A thorough investigation indicates the coordinate-based perturbation performs worse because of the point sampling (elaborated in Section~\ref{subsec:dl}) by the PCSS model. For example, when a point cloud is fed into ResGCN-28, it is sampled by a $k$-NN algorithm that aggregates the neighborhood points to the centric points. When the coordinates of a point are perturbed, the nearby points are also changed due to point aggregation, so the result of the attack might not be controllable. As a piece of supporting evidence, we sampled the whole point clouds on Area 5 from the S3IDS dataset and found \textit{over 88\%} of the neighborhood points are changed after coordinate-based perturbation. 
Perturbing both coordinate and color also leads to unsatisfactory results for the same reason. 
On the other hand, the perturbing color will not impact how neighborhood points will be sampled, so the attack outcome is more controllable.

Given that perturbing color is much more effective than perturbing coordinates, for the following experiments, all attacks are conducted under color-based perturbation, and we switch the distance back to the default $L_2$.


\begin{tcolorbox}[colback=white, left=0pt, right=0pt, top=0pt, bottom=0pt]
\noindent \textbf{Finding 1}:  
The color feature is more vulnerable than point coordinates under perturbation.
\end{tcolorbox}

\ignore{
Our result in Table~\ref{tab:nt-l0-performance}  shows that color-based perturbation outperforms coordinate-based perturbation in the degradation of segmentation accuracy and aIoU (they are only counted on the successful samples). Moreover, the sample success rate (SSR) is significantly higher (81.17\% versus 11.11\% for ResGCN-28). In other words, \textbf{the color features are more vulnerable}.
}

\subsection{Evaluation on Performance Degradation Attack}
\label{subsec:non_targeted}

\begin{table*}[ht]
    \caption{Performance degradation attack against PointNet++, ResGCN-28, and RandLA-Net on S3DIS. The color feature is attacked and $L_2$ distance is used. $\downarrow$ shows the drop of PCSS accuracy or aIoU.
    }
    \centering
    \begin{tabular}{ll|lll|lll|lll}
    
    \hline 

    \multirow{2}{*}{}& & \multicolumn{3}{c|}{Random Noise}  &  \multicolumn{3}{c}{Norm-unbounded}  &  \multicolumn{3}{|c}{Norm-bounded}\\     

    \textbf{Case} & & {$\mathbf{L_2}$} & \textbf{Acc} (\%) & \textbf{aIoU} (\%) & $\mathbf{L_2}$ & \textbf{Acc} (\%) & \textbf{aIoU} (\%) & $\mathbf{L_2}$ & \textbf{Acc} (\%) & \textbf{aIoU} (\%) \\
    \hline
    \hline
    \multirow{3}{*}{PointNet++} & Best    & 2.68 & 14.51 & 14.26  & 2.68 & 4.91 & 2.31 & 15.51 & 5.71 & 5.19 \\

    & Avg & 18.19 & 77.26(5.39$\downarrow$) & 70.02(0.69$\downarrow$) & 18.27 & \textbf{7.86}(74.79$\downarrow$) & \textbf{8.85}(61.86$\downarrow$) & 18.27 &19.11(63.54$\downarrow$) & 16.10(54.61$\downarrow$) \\
    & Worst   & 30.02 & 100 & 100 & 30.03 & 20.33 & 59.27 & 22.01 & 56.71 & 42.37 \\
    \hline

    \multirow{3}{*}{ResGCN-28} & Best    & 1.29 & 7.79 & 4.11  & 1.29 & 0.31 & 0.16 & 5.05 & 0.0 & 0.0 \\
    & Avg & 4.30 & 82.36(3.54$\downarrow$) & 73.17(6.55$\downarrow$) & 4.30 & \textbf{6.75}(79.15$\downarrow$) & \textbf{3.49}(76.23$\downarrow$) & 6.51 & 42.16(43.74$\downarrow$) & 30.13(49.59$\downarrow$) \\
    & Worst   & 9.81 & 100 & 100 & 9.81 & 18.34 & 10.10 & 7.96 & 99.85 & 99.70 \\
    \hline
    
    \multirow{3}{*}{RandLA-Net} & Best    & 6.65  & 34.07 & 12.56  & 6.65 & 6.13 & 0.92 & 0.33 & 18.52 & 13.62 \\
    & Avg & 17.06 & 78.01(6.24$\downarrow$) & 44.82(6.6$\downarrow$) & 17.06 & \textbf{7.45}(76.75$\downarrow$) & \textbf{2.96}(48.45$\downarrow$) & 16.83 & 59.60(24.65$\downarrow$) & 31.15(20.26$\downarrow$)  \\
    & Worst & 54.62 & 97.18 & 70.08  & 54.62 & 7.69 & 8.33 & 17.00 & 85.88 & 66.18 \\
    \hline
    \end{tabular}

    \label{tab:nt-s3dis-performance}
\end{table*}



\ignore{
\begin{tcolorbox}[colback=white, left=0pt, right=0pt, top=0pt, bottom=0pt]
\noindent 
\textbf{RQ2:} Which attack method is more effective under performance degradation attack? \jxu{performance attack} \\
\textbf{RQ3:} Does attack effectiveness varies by the target model?
\end{tcolorbox}
}

In this subsection, we conduct experiments under performance degradation attack, focusing on the attack effectiveness under different methods and target models. We use S3IDS as the dataset.
For the attack methods, in addition to norm-unbounded and norm-bounded attacks, we also implement another method that adds random noises to the color channels, as a baseline to compare against. 

In Table~\ref{tab:nt-s3dis-performance}, we show accuracy and aIoU across the point clouds. 
We also show the $L_2$ distance between the original point cloud and the perturbed one.
It turns out norm-unbounded and norm-bounded attacks both significantly reduce the accuracy and aIoU. For example, norm-unbounded attack \textit{drops the average accuracy of PointNet++, ResGCN-28, and RandLA-Net from 82.65\% to 7.86\%, 85.90\% to 6.75\%, and 87.2\% to 7.45\% separately}. The average drop rate of aIoU ranges from 48.45\% to 76.23\% under norm-unbounded attack, and from 20.26 to 54.61\% under norm-bounded attack.
Norm-unbounded attack performs much better for the worst-case scenario (the most difficult sample): e.g., when ResGCN-28 is attacked, the accuracy on the most difficult sample is dropped to 18.34\%, but the norm-bounded attack has nearly no impacts on the accuracy (99.85\%).

Regarding the perturbation distance, we found that norm-unbounded attack generates the adversarial examples under smaller or equal distance in the best-case scenario and average scenarios for PointNet++ and ResGCN-28, but the distance becomes larger for RandLA-Net (e.g., 17.06 compared to 16.83 for the average). For the worst-case scenario, it has to add much larger noises to drop accuracy and aIoU.


Regarding the baseline method, we found its effectiveness is quite limited, with the dropping of average accuracy and aIoU ranging from 3.54\% to 6.24\% and 0.69\% to 6.6\% respectively. The result suggests PCSS models are robust against random noises and carrying out a successful attack is non-trivial. 

Regarding the impact of the target model on the attack effectiveness, we found norm-unbounded attack is similarly effective against PointNet++, ResGCN-28, and RandLA-Net, but the norm-bounded attack is much more effective against PointNet++ than ResGCN-28 and RandLA-Net.
Hence, we suggest that the norm-unbounded attack should be used if the attacker prefers effectiveness, or finding better adversarial examples. In the meantime, it usually takes a longer time to execute due to the longer time to reach the converge requirement. 

\revised{Finally, we measure the overhead of conducting the attacks.
The time to generate an adversarial example is proportional to the number of Steps (see Algorithm~\ref{alg:algorithm_pgd}), and each step takes 0.3 seconds for the norm-bounded attack, and 0.2 for the norm-unbounded attack. 
This overhead is acceptable if the attacker uses physical patches~\cite{tu2020physically} or adversarial objects~\cite{zhu2021adversarial}, as they are generated offline. 
The overhead can be further reduced by using a smaller number of Steps. Recently, Guesmi et al. showed it is possible to conduct real-time adversarial attacks by introducing an offline component~\cite{guesmi2022room}, and our attacks can be adjusted following this direction.
\ignore{
The cost times for single point clouds are based on the paramters of the attacks and the robustness of the point cloud. 
In the experiment, to achieve high attack success rate, we use more iteration and the time cost for each iteration is around 0.05 seconds as point cloud perturbation in our experiment environment. For the row object preturbation, we can update the attack parameters based on the attack target, hardware calculating time and singal processing time to meet the real-time attack requirements.
}
}

\begin{tcolorbox}[colback=white, left=0pt, right=0pt, top=0pt, bottom=0pt]
\noindent \textbf{Finding 2}:  
Under performance degradation attack,  the norm-unbounded attack is more effective, especially for the most difficult point cloud samples.
\\
\noindent \textbf{Finding 3}:  
All tested models are similarly vulnerable under norm-unbounded attack, but PointNet++ is much more vulnerable under norm-bounded attack.
\end{tcolorbox}

\subsection{Evaluation on Object Hiding Attack}
\label{subsec:targeted}

\begin{table}
    \caption{The results of norm-unbounded attack on window (label=5), door (label=6), table (label=7), chair (label=8), bookcase (label=10), board (label=11). ``PN'' means PointNet++, ``RGCN'' means ResGCN-28, ``RNet'' means RandLA-Net, ``SC'' means Source Class. ``OOB/Acc'' means the out-of-band accuracy and overall accuracy while  ``OOB/aIoU'' is similar.}
    \centering
    \begin{tabular}{llcccc}
        \hline

        \textbf{Model} & \textbf{SC} & $\mathbf{L}_2$ & \textbf{PSR (\%)} & \textbf{OOB/Acc (\%)} & \textbf{OOB/aIoU (\%)}  \\
        \hline
        \hline

        \multirow{6}{*}{PN} & 5 & 7.67 & \textbf{93.92} & 53.76 / 77.67 & 46.59 / 60.77 \\
        
        & 6 & 5.39 & \textbf{93.11} & 58.54 / 62.00 & 45.37 / 49.24 \\
        
        & 7 & 10.55 & 37.70 & 79.48 / 86.26 & 56.04 / 69.09 \\
        
        & 8 & 6.69 & 17.63 & 86.09 / 90.65 & 62.91 / 73.19 \\
        & 10 & 15.26 & \textbf{93.25} & 52.73 / 68.43 & 49.63 / 57.88 \\
        & 11 & 5.28 & \textbf{93.16} & 80.47 / 93.97 & 60.99 / 74.27 \\
        \hline        

        \multirow{6}{*}{RGCN} & 5 & 14.57 & \textbf{95.44} & 70.88 / 71.21 & 39.57 / 58.67 \\
        & 6 & 12.17 & \textbf{94.71} & 77.96 / 84.62 & 65.11 / 76.75 \\
        & 7 & 9.29 & 66.43 & 81.81 / 91.66 & 49.08 / 84.80 \\
        & 8 & 12.62  & 51.63 & 82.22 / 83.84 & 63.39 / 75.59  \\

        & 10 & 16.01 & \textbf{90.48} & 65.10 / 68.43  & 55.56 / 56.52  \\
        & 11 & 9.69 & \textbf{96.08} & 78.37 / 88.85 & 56.58 / 66.43 \\
        \hline

        \multirow{6}{*}{RNet} & 5 & 3.76 & \textbf{95.13} & 83.98 / 84.41 & 50.39 / 51.03 \\
        & 6 & 2.79 & \textbf{95.23} & 88.42 / 88.78 & 49.52 / 51.12 \\
        & 7 & 11.82 & 83.10 & 83.11 / 83.98 & 45.01 / 50.58 \\
        & 8 & 9.06 & 85.98 & 82.78 / 82.94 & 47.50 / 47.94 \\
        & 10 & 8.55 & \textbf{95.05} & 84.02 / 84.71 & 50.07 / 51.27 \\
        & 11 & 2.37 & \textbf{94.29} & 84.80 / 85.57 & 52.22 / 54.06 \\
        \hline  
    \end{tabular}

    \label{tab:t-performance}
\end{table}

\begin{table}
    \caption{The results of norm-bounded attack on window (label=5), door (label=6), table (label=7), chair (label=8), bookcase (label=10), board (label=11). ``PN'' means PointNet++, ``RGCN'' means ResGCN-28, ``RNet'' means RandLA-Net, ``SC'' means Source Class.``OOB/Acc'' means the out-of-band accuracy and overall accuracy while  ``OOB/aIoU'' is similar.}
    \centering
    \begin{tabular}{llcccc}
    
        \hline

        \textbf{Model} & \textbf{SC} & $\mathbf{L}_2$ & \textbf{PSR (\%)} & \textbf{OOB/Acc (\%)} & \textbf{OOB/aIoU (\%)}  \\
        \hline
        \hline

        \multirow{6}{*}{PN} & 5 & 5.44 & 81.12 & 34.55 / 81.66 & 32.53 / 70.31 \\
        & 6 & 3.78 & 42.85 & 88.72 / 94.67 & 52.42 / 66.60 \\
        & 7 & 5.78 & 3.84 & 60.71 / 85.67 & 52.42 / 67.20 \\
        & 8 & 3.02 & 1.04 & 65.33 / 85.24 & 42.25 / 67.20 \\
        & 10 & 5.26 & 42.22 & 47.60 / 71.44 & 41.24 / 61.49 \\
        & 11 & 6.85 & 70.58 & 74.55 / 89.23 & 48.37 / 63.41 \\
        \hline        

        \multirow{6}{*}{RGCN} & 5 & 4.25 & 65.80 & 44.60 / 80.90 & 42.53 / 69.31 \\
        & 6 & 4.16 & 26.27 & 65.76 / 88.03 & 65.72 / 79.88 \\
        & 7 & 4.23 & 1.24 & 63.89 / 88.85 & 61.24 / 81.73 \\
        & 8 & 4.35 & 0.90 & 62.00 / 93.02 & 59.73 / 88.02  \\
        & 10 & 5.87 & 7.35 & 45.46 / 83.72  & 58.00 / 73.93  \\
        & 11 & 3.83 & 26.20 & 67.59 / 91.15 & 64.87 / 84.67 \\
        \hline       
        
        \multirow{6}{*}{RNet} & 5 & 3.86 & 82.42 & 81.50 / 84.42 & 44.93 / 50.48 \\
        & 6 & 3.97 & 83.05 & 81.82 / 84.57 & 44.28 / 50.18 \\
        & 7 & 4.42 & 80.95 & 80.85 / 84.55 & 44.14 / 51.20 \\
        & 8 & 4.19 & 83.59 & 81.55 / 83.96 & 45.31 / 50.95 \\
        & 10 & 3.99 & 91.87 & 82.88 / 84.66 & 47.38 / 51.28 \\
        & 11 & 2.83 & 93.95 & 86.45 / 86.74 & 52.73 / 56.69 \\
        \hline  
    \end{tabular}

    \label{tab:t-performance-pgd}
\end{table}

We used Area 5 of S3IDS for evaluation, which contains objects under 13 classes, and we perturb the points from 6 classes, including window (label=5), door (label=6), table (label=7), chair (label=8), bookcase (label=10), and board (label=11), because the quantity of points of each class is not too small. We set the target class as wall (label=2). 

Table~\ref{tab:t-performance} shows the results for the 6 objects under norm-unbounded attack. It turns out \textit{PSR can be over 90\%} for window, door, bookcase, and board, against all targeted models. However, PSR is much lower for table and chair.
We speculate the reason is that table and chair have more complex shapes, 
\ignore{and the similarity of the object shape between the table and the wall is huge.} so changing the class labels on these objects is more difficult.
In the meantime, the drop in the accuracy of the OOB points is moderate, mostly within 10\%, which suggests norn-unbounded is able to confine the changes to the selected objects.

Table~\ref{tab:t-performance-pgd} shows the results under norm-bounded attack. Similar to the trend of the performance degradation attack, it performs worse than the norm-unbounded attack, i.e., lower PSR for every perturbed object. Regarding the impact of objects, we also found PSR is higher when less complex objects like window, door, bookcase and board are perturbed, but table and chair see much lower PSR, e.g., under 10\% for PointNet++ and ResGCN-28. Moreover, norm-bounded attack results in a larger drop of OOB accuracy and aIOU, especially for PointNet++ and ResGCN-28.
Since the design of norm-bounded attack bounds the perturbation distance, we found the $L_2$ distance of adversarial samples is much smaller in most cases, though it comes at the price of lower PSR.

Regarding the target model, we found it is easier to achieve high PSR when attacking RandLA-Net. For example, even table and chair have over 80\% PSR for both attack methods. 

\begin{tcolorbox}[colback=white, left=0pt, right=0pt, top=0pt, bottom=0pt]
\noindent 
\textbf{Finding 4:}  Under object hiding attack, norm-unbounded attack is also  more effective.  \\
\textbf{Finding 5:}  Source class has big impact on the attack effectiveness, as changing the labels on the complex objects is much harder.
\end{tcolorbox}

\subsection{Evaluation on Outdoor Dataset}
\label{outdoor}

\begin{table*}[t]
    \caption{The results of the Performance degradation attack from RandLA-Net on Semantic3D.}
    
    \centering
    \begin{tabular}{l|lll|lll}

    \hline 
    \multirow{2}{*}{\textbf{Case}} &  \multicolumn{3}{c|}{Random Noise} & \multicolumn{3}{c}{Norm-unbounded} \\ 

    & {$\mathbf{L_2}$} & \textbf{Acc} (\%) & \textbf{aIoU} (\%) & $\mathbf{L_2}$ & \textbf{Acc} (\%) & \textbf{aIoU} (\%)  \\

    \hline
    \hline
    Best    & 19.31  & 0.00 & 0.00  & 19.31 & 0.00 & 0.00 \\
    Average & 25.84 & 79.30(18.95$\downarrow$) & 37.22(26.04 $\downarrow$) & 25.84 & \textbf{16.00}(82.25$\downarrow$) & \textbf{7.70}(55.56\%$\downarrow$) \\
    Worst   & 280.79 & 100.0 & 100.0  & 280.79 & 90.82 & 25.42 \\
    \hline
    
    \end{tabular}
    \label{tab:nt-randlanet-semantic3d-performance}
\end{table*}

\begin{table*}[h]
    \caption{The results of the object hiding attack against RandLA-Net on Semantic3d. Car (label=8) is perturbed to man-made terrain (label=1), natural terrain (label=2), high vegetation (label=3), low vegetation (label=4).
    }
    \centering
    \begin{tabular}{lllll}
    
        \hline
        \textbf{Target Class} & $\mathbf{L}_2$ &\textbf{PSR}&\textbf{OOB Acc /Acc}(\%)&\textbf{OOB aIoU /aIoU}(\%)\\
        \hline
        \hline
        
        Man-made terrain & 10.41 & 85.30 & 73.03 / 73.64  & 30.74 / 33.56 \\
        \hline        
        Natural terrain & 5.61 & 73.96 & 84.76 / 84.89 & 46.63 / 48.19 \\
        \hline
        High vegetation & 3.61  & \textbf{94.26} & 97.95 / 97.99 & 58.86 / 61.51 \\
        \hline
        Low vegetation & 3.60  & \textbf{94.86} & 74.18 / 74.70 & 39.57 / 42.52 \\
        \hline  
    \end{tabular}

    \label{tab:t-semantic3d-performance}
\end{table*}

        

\ignore{
\begin{tcolorbox}[colback=white, left=0pt, right=0pt, top=0pt, bottom=0pt]
\noindent 
\textbf{RQ6:} Is outdoor scene also vulnerable?
\end{tcolorbox}
}

All previous evaluations are carried out on S3IDS, an indoor dataset. Since the outdoor scenes have different properties (e.g., different object classes and larger sizes), we evaluate the attacks against another outdoor dataset, Semantic3D. Only RandLA-Net is attacked because the other two PCSS models cannot handle the scale of point clouds in Semantic3D. We show the result of norm-unbounded attack only as it is more effective in previous experiments.

Table~\ref{tab:nt-randlanet-semantic3d-performance} shows the results of the performance degradation attack. Similar to the attack against S3IDS, the norm-unbounded attack greatly decreases the accuracy and aIoU compared to the baseline (random noises) when they target the same $L_2$ distance: the average accuracy and the aIoU drop from 98.25\% and 63.26\% to 16.00\% and 7.70\%. The baseline only drops the average accuracy and the aIoU to 79.30\% and 37.22\%. 
We also observe that the result on RandLA-Net has a bigger variance by samples: e.g., the accuracy can drop to 0\% for the best case, and 90.82\% for the worst case.

As for the object hiding attack, we manipulate the source points labeled as car (label=8) to mislead the model to predict them as man-made terrain (label=1), natural terrain (label=2), high vegetation (label=3) and low vegetation (label=4). From Table~\ref{tab:t-semantic3d-performance}, PSR is near 95\% when vegetation is the target class. Though the outdoor scene is expected to be more complex, our result shows object hiding attack is still effective. 

\begin{tcolorbox}[colback=white, left=0pt, right=0pt, top=0pt, bottom=0pt]
\noindent 
\textbf{Finding 6:}  Norm-unbounded attack is also effective when attacking an outdoor scene, under both the performance degradation and the object hiding attacks.
\end{tcolorbox}

\subsection{Defense Methods}
\label{subsec:defense}

To mitigate the threats from the proposed adversarial attacks, gradient obfuscation, adversarial training, and anomaly detection can be tested on PCSS. These ideas have been initially examined on 2D image classification and were recently migrated to 3D point cloud object recognition. For gradient obfuscation, DUP-Net~\cite{zhou2019dup} includes a denoiser layer and upsampler network structure. GvG-PointNet++~\cite{dong2020self} introduces gather vectors in the points clouds. 
Recently, Li et al.~\cite{li2022robust} proposed implicit gradients, which could lead the attackers to the wrong updating directions, to replace obfuscated gradients. 
For adversarial training, DeepSym~\cite{sun2020adversarial} uses a sorting-based pooling operation to overcome the issues caused by the default symmetric function. \revised{Sun et al.~\cite{sun2021adversarially} analyze the robustness of self-supervised learning 3D point cloud models with adversarial training.}
For anomaly detection, Yang et al.~\cite{yang2019adversarial} and Rusu et al. ~\cite{rusu2008towards} measured the statistics of point clouds to detect or mitigate the attacks.


\revised{Here, we measure how our attacks are impacted when the defenses are deployed, and we select the approaches under anomaly detection, as they are lightweight (e.g., adversarial training is heavyweight because it incurs high training overhead). We select two defense methods: Simple Random Sampling (SRS)~\cite{yang2019adversarial} and Statistical Outlier Removal (SOR)\footnote{We use the code from \url{https://github.com/Wuziyi616/IF-Defense}.}~\cite{zhou2019dup}. More specifically, SRS filters out a subset of points from a point cloud to mitigate the impact of perturbations while SOR removes the outlier points based on a $k$-NN distance. 
\ignore{We do not consider training-based method like DUP-Net~\cite{zhou2019dup}, IF-Defense~\cite{wu2020if}, GvG-PointNet++~\cite{dong2020self} because these methods are designed to work on point cloud single object detection task and the trade-off between robustness and accuracy is uncertain on the semantic segmentation task.} 
SRS can be directly used in semantic segmentation. For SOR, we revise the $k$-NN distance function by using both color and coordinate. The sampling number of SRS is 50 (around 1\% of the whole point cloud number) and $k$ is 2 for SOR. We follow the experiment setting from IF-Defense~\cite{wu2020if} and randomly select 100 point clouds to make a fair comparison. We evaluate our attacks on S3DIS with ResGCN-28 as the PCSS model.}

\begin{table}[h]
    \caption{The ResGCN-28 result with SRS and SOR under performance degradation attack.}
    \centering
    \begin{tabular}{lllll}
        \hline
        \textbf{Attack} & \textbf{Defense} & $L_2$ & \textbf{Acc} (\%) & \textbf{aIoU} (\%)  \\
        \hline
        \hline
        \multirow{3}{*}{Norm-bounded} & None & 6.50 & 42.06 & 30.13 \\
        & SRS & 6.57 & 46.06 & 36.18  \\
        & SOR & 6.56 & 46.88 & 36.92 \\
        \hline
        \multirow{3}{*}{Norm-unbounded} & None & 4.30 & 6.85 & 3.79 \\
        & SRS & 6.22 & 10.54 & 5.70  \\
        & SOR & 9.44 & 41.48 & 29.86  \\
        \hline
    \end{tabular}
    \label{tab:defense}
\end{table}
\shephred{
As the results in Table~\ref{tab:defense} show, the norm-bounded attack is still effective even when the two defense methods are applied (similar accuracy and aIoU with or without defenses). For the norm-unbounded attack, 
as the $L_2$ distance cannot be fixed to a value, we try different attack parameters to make the $L_2$ distance fall in a similar range.
It turns out in this case, the changes are more likely considered as outlier and detected by SOR (SOR reaches higher accuracy than SRS). Still, none of the defenses are able to restore the accuracy and aIoU to the original levels, i.e., over 70\%.
A similar observation was also made in ~\cite{wu2020if}.
}

\begin{tcolorbox}[colback=white, left=0pt, right=0pt, top=0pt, bottom=0pt]
\noindent 
\textbf{Finding 7:}  
The defenses based on anomaly detection are ineffective against norm-bounded attacks. Norm-unbounded attacks are affected more under Statistical Outlier Removal (SOR).
\end{tcolorbox}

\subsection{Attack Transferability}
\label{transferability}

\begin{table}[h]
    \caption{The upper row shows the results of attack transferability on PointNet++. The lower row displays the results from transferring ResGCN-28 adversarial samples to PointNet++.}
    \centering
    \begin{tabular}{lll}
    
        \hline
        \textbf{PCSS Model} & \textbf{Acc} (\%) & \textbf{aIoU} (\%)  \\
        \hline
        \hline
        PointNet++ (Pre-trained) & 7.24 & 9.44 \\
        PointNet++ (Self-trained) & 34.35 & 31.39 \\
        \hline
        
        ResGCN-28 & 7.11 & 3.68 \\
        PointNet++ & 39.01 & 25.30  \\    
        \hline
    \end{tabular}

    \label{tab:nt-tran-similar-performance}
\end{table}

\ignore{
\begin{tcolorbox}[colback=white, left=0pt, right=0pt, top=0pt, bottom=0pt]
\noindent 
\textbf{RQ7:} Can the generated adversarial example be applied on another PCSS model?
\end{tcolorbox}
}

Existing research has shown an adversarial sample targeting 2D image classification has transferability~\cite{papernot2016transferability}, i.e., that a sample generated against one model is also effective against another model.
We are interested in whether our adversarial samples have the same property. To this end, we first evaluate the attack transferability on models with different parameters.
We select 400 adversarial samples generated by the norm-unbounded attack on the pre-trained PointNet++, and feed them into another PointNet++ trained by ourselves (the weights and biases are different).
The results in Table~\ref{tab:nt-tran-similar-performance} show the accuracy and aIoU on the 400 samples, and both are less than 40\%, suggesting our adversarial samples are transferable under different model parameters.


Then we test transferability across model families: we generate adversarial examples for ResGCN-28 and test if they can fool PointNet++. 
We do not transfer the attack against RandLA-Net due to its highly different approach of pre-processing. 
Due to different normalization steps (i.e., the coordinate ranges in $[-1,1]$ for ResGCN-28 while $[0,3]$ for PointNet++), the adversarial samples do not directly transfer, so we perform an extra step to map the attacked fields to the same range.
\ignore{with Equation \ref{eq:res-point} before they are fed into PointNet++. \zl{why do we change coordinates if we attack color?}
\begin{equation}
    p_i = \{2* pos_x, 2*pos_y, \frac{3}{2}*pos_z+\frac{3}{2}\}
    \label{eq:res-point}
\end{equation}
}
We compute the accuracy and aIoU in the two settings, and the results suggest our adversarial examples are also transferrable (Table~\ref{tab:nt-tran-similar-performance}).

\ignore{Interestingly, Xiang et al.~\cite{xiang2019adv} showed transferability is limited no matter if the same model with different parameters or different models are tested. Our result on semantic segmentation draws a different conclusion.
}

\begin{tcolorbox}[colback=white, left=0pt, right=0pt, top=0pt, bottom=0pt]
\noindent 
\textbf{Finding 8:}  The adversarial example is transferable under different model parameters and across model families.
\end{tcolorbox}



\subsection{Visualization of Adversarial Examples}
\label{subsec:visualization}


\ignore{
\begin{figure}[h]
     \includegraphics[width=0.49\columnwidth]{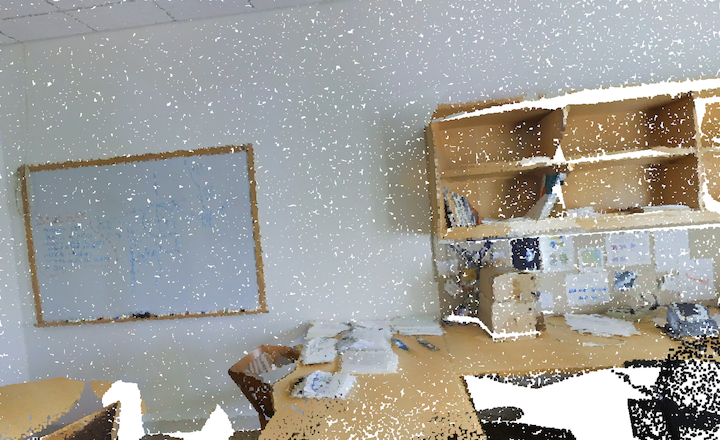}\hfill
     \includegraphics[width=0.49\columnwidth]{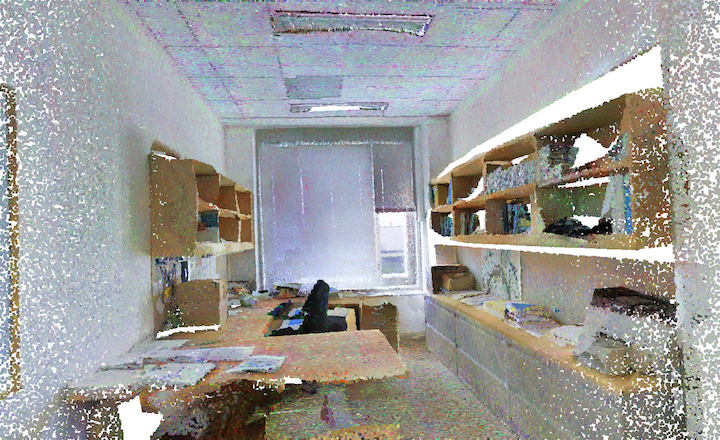}\hfill \\
     \includegraphics[width=0.49\columnwidth]{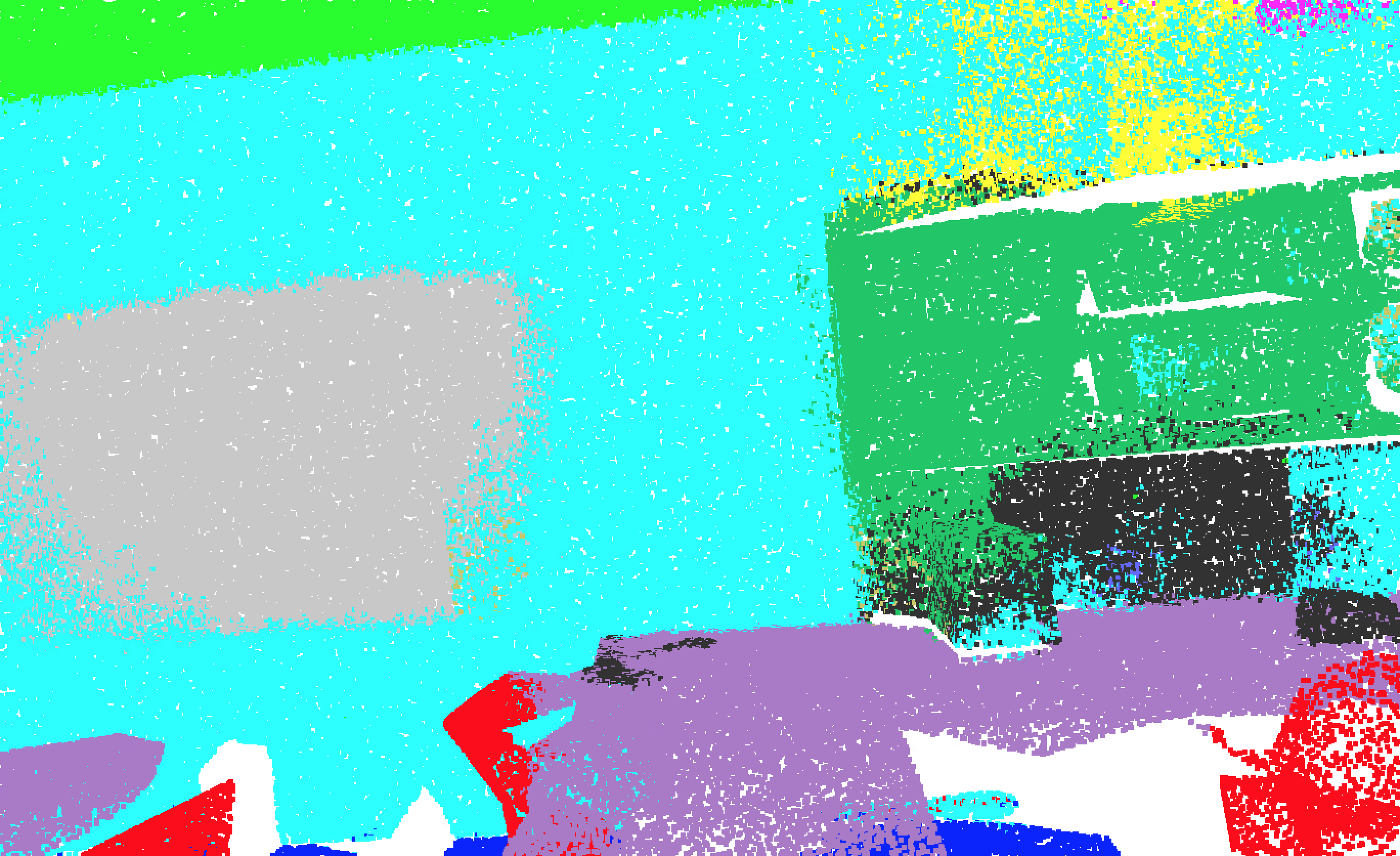}\hfill
     \includegraphics[width=0.49\columnwidth]{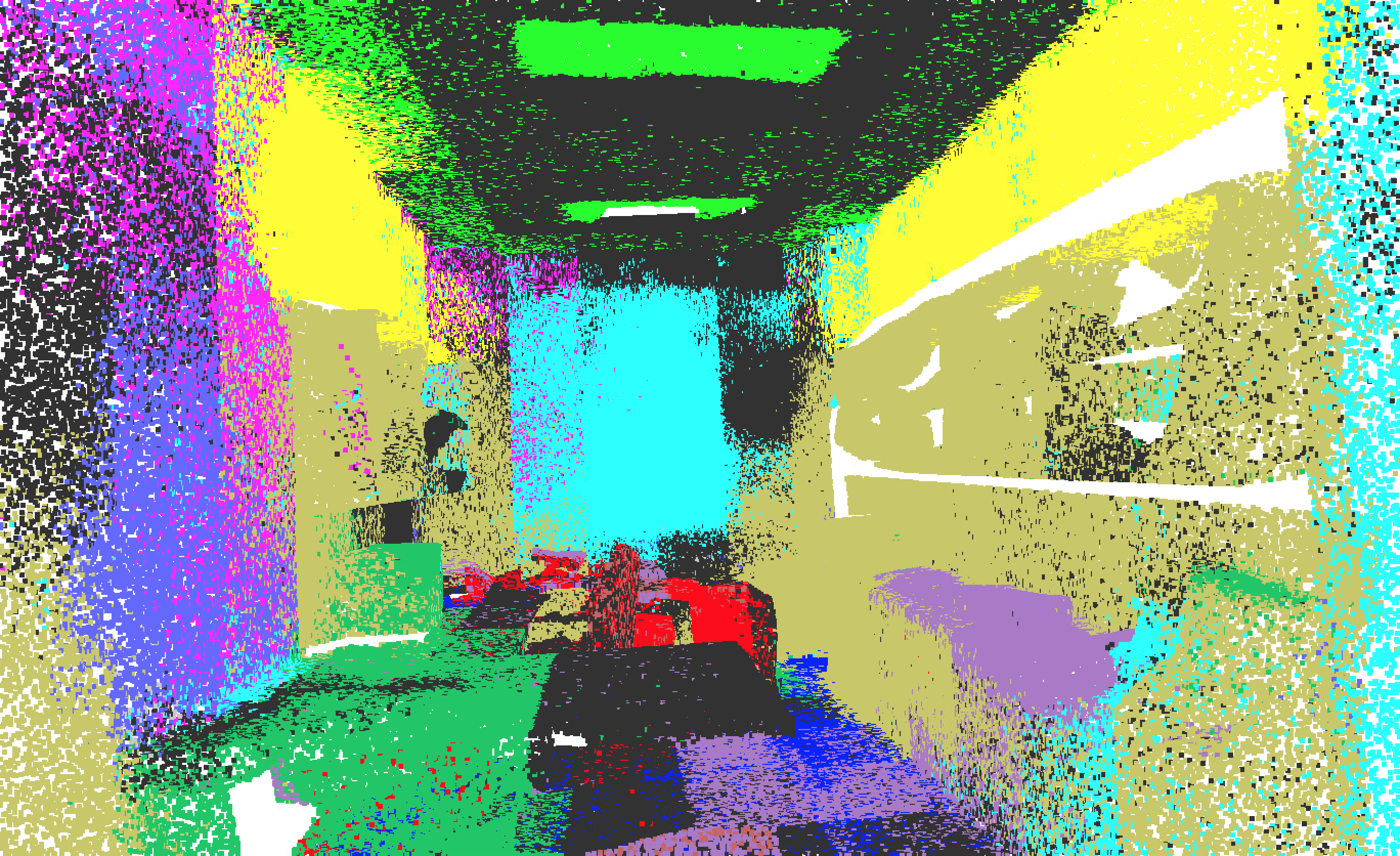}\hfill
    \caption{The upper figures show the perturbed scene of Office 33 of Area 5 in S3DIS. The lower figures display the segmentation result of the perturbed point clouds. \zl{commented it out because it doesn't compare before and after perturbation}
    }\label{Fig:target_example_s3dis}
\end{figure}
}

\ignore{
As described in Section~\ref{sec:problem}, the attacker's goal is to change the prediction results of points in $X_T$. Here we visualize the attack outcome with one example, and we assume PointNet++ is the target model. From Figure~\ref{Fig:target_example_s3dis}, The original scene of Office 33 of Area 5 is shown in the upper left and the lower left one is its corresponding segmentation result.  We consider $X_t$ as all the points in the point clouds and apply \Mname. The segmentation result is changed to the lower right figure.
For instance, the points related to the desk at the bottom left are misclassified as a bookcase. 
As the two upper figures from Figure~\ref{Fig:target_example_s3dis} show, their difference is insignificant. 
}

\begin{figure*}[h]
    \includegraphics[width=\textwidth]{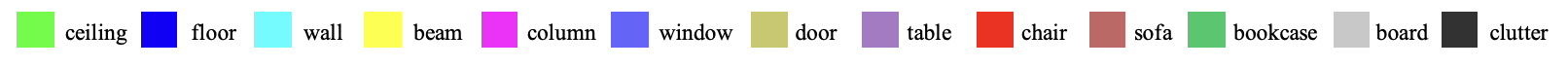} \\

    \includegraphics[width=0.24\textwidth]{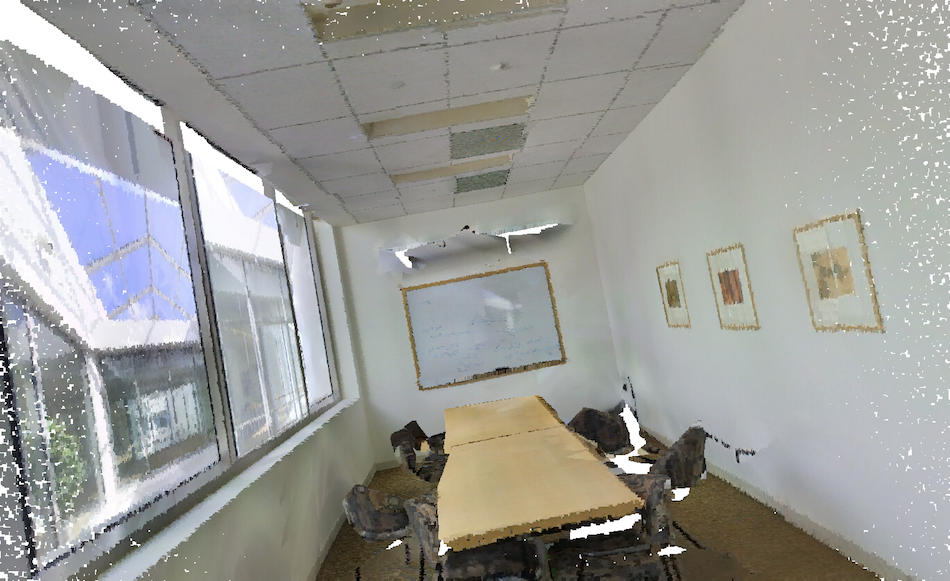} \hfill
    \includegraphics[width=0.24\textwidth]{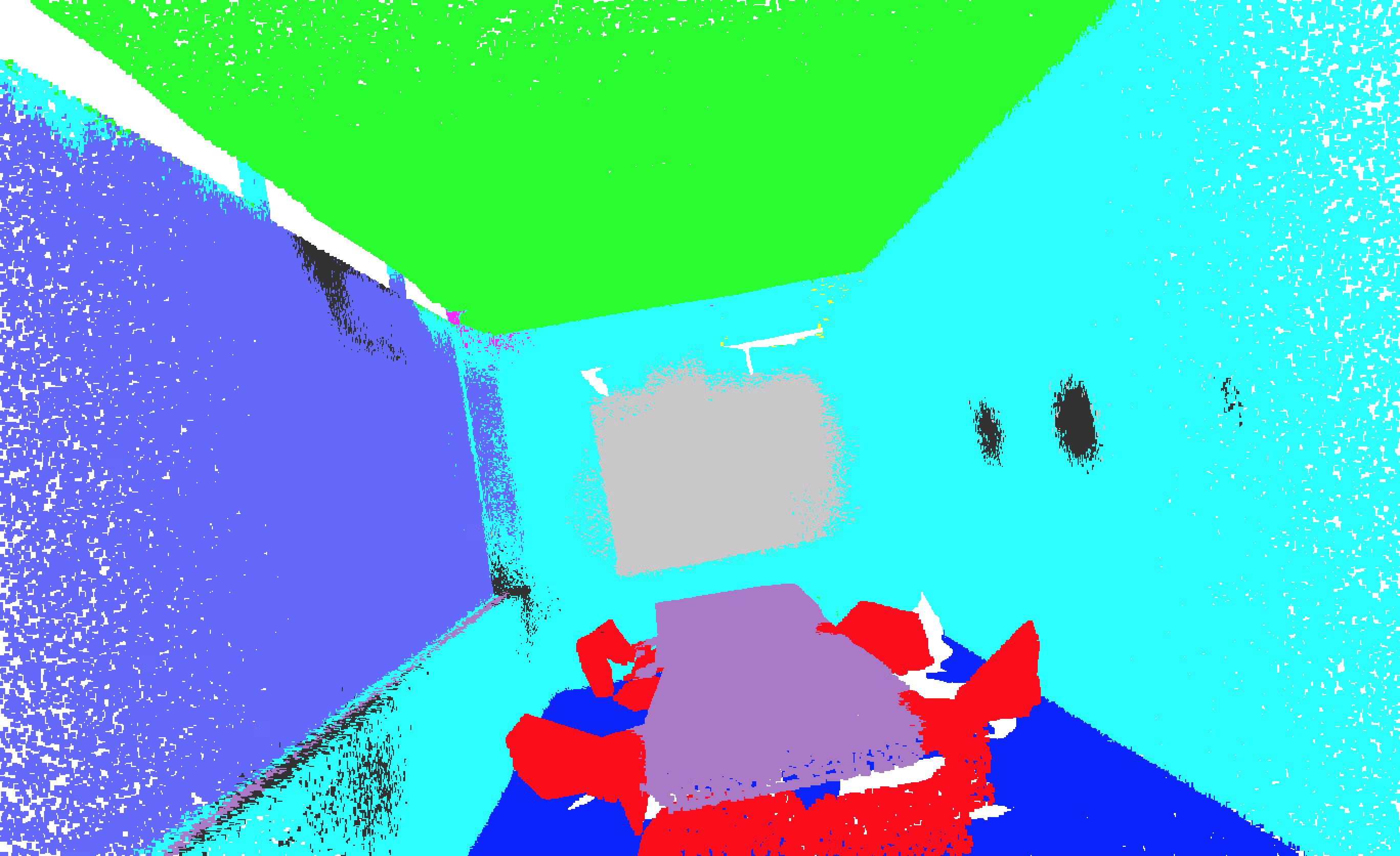}\hfill
    \includegraphics[width=0.24\textwidth]{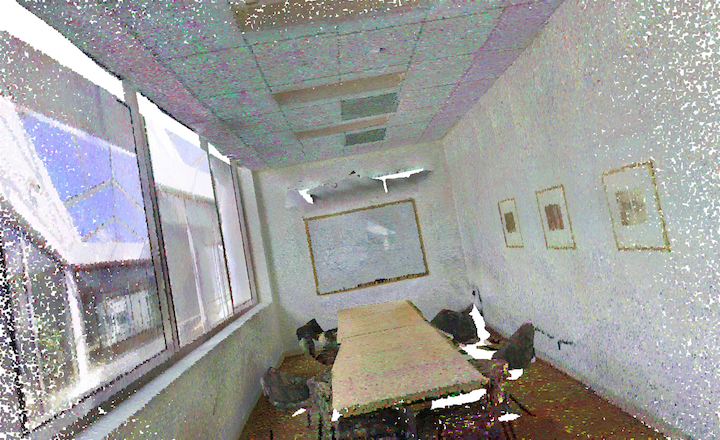}\hfill 
    \includegraphics[width=0.24\textwidth]{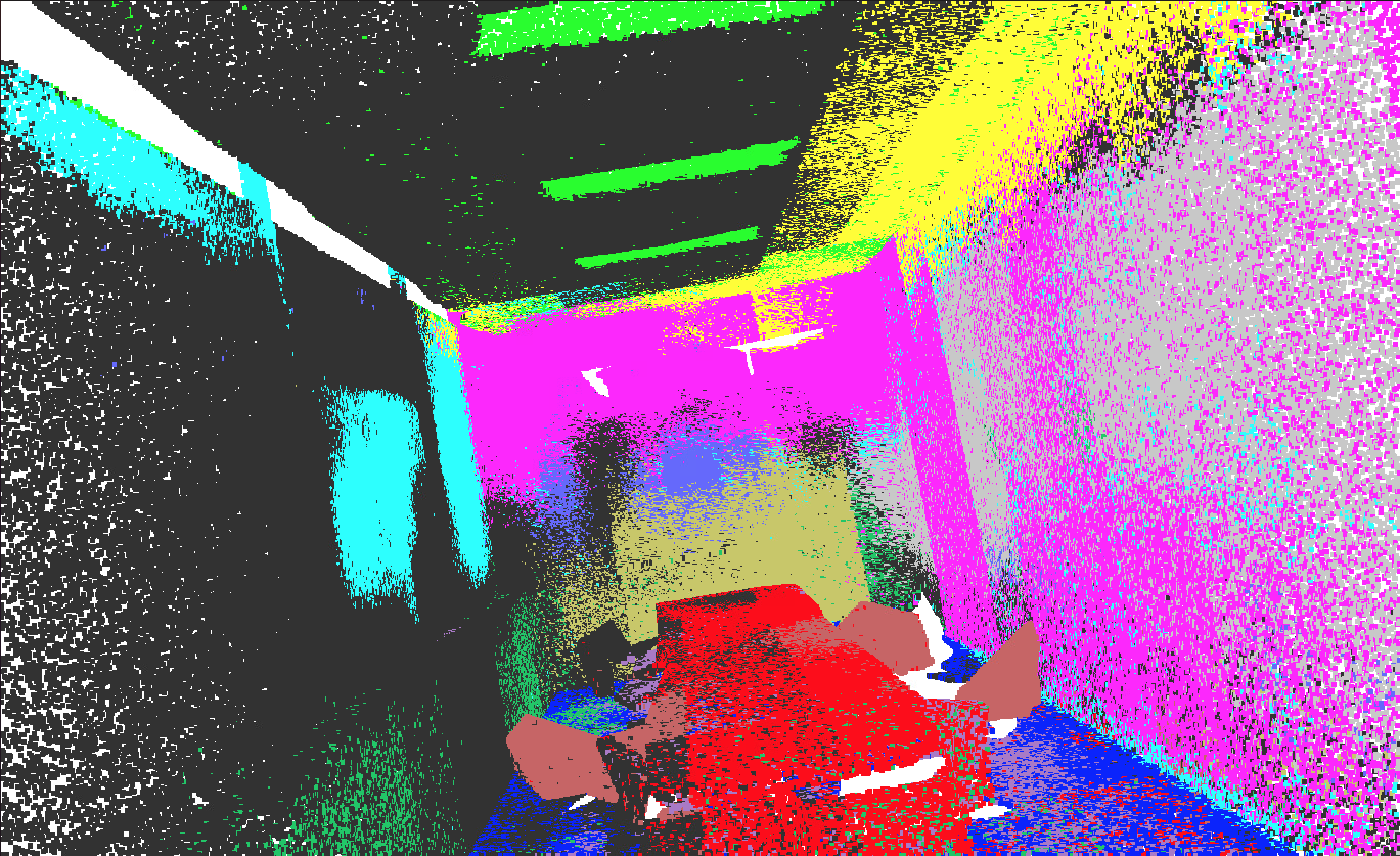}\hfill
    
    \includegraphics[width=0.24\textwidth]{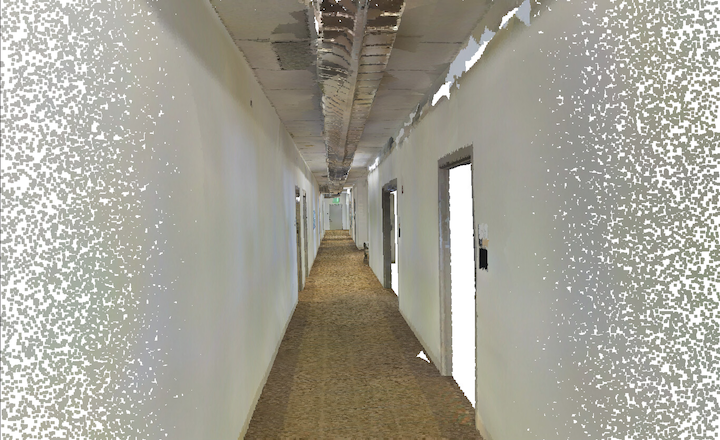}\hfill
    \includegraphics[width=0.24\textwidth]{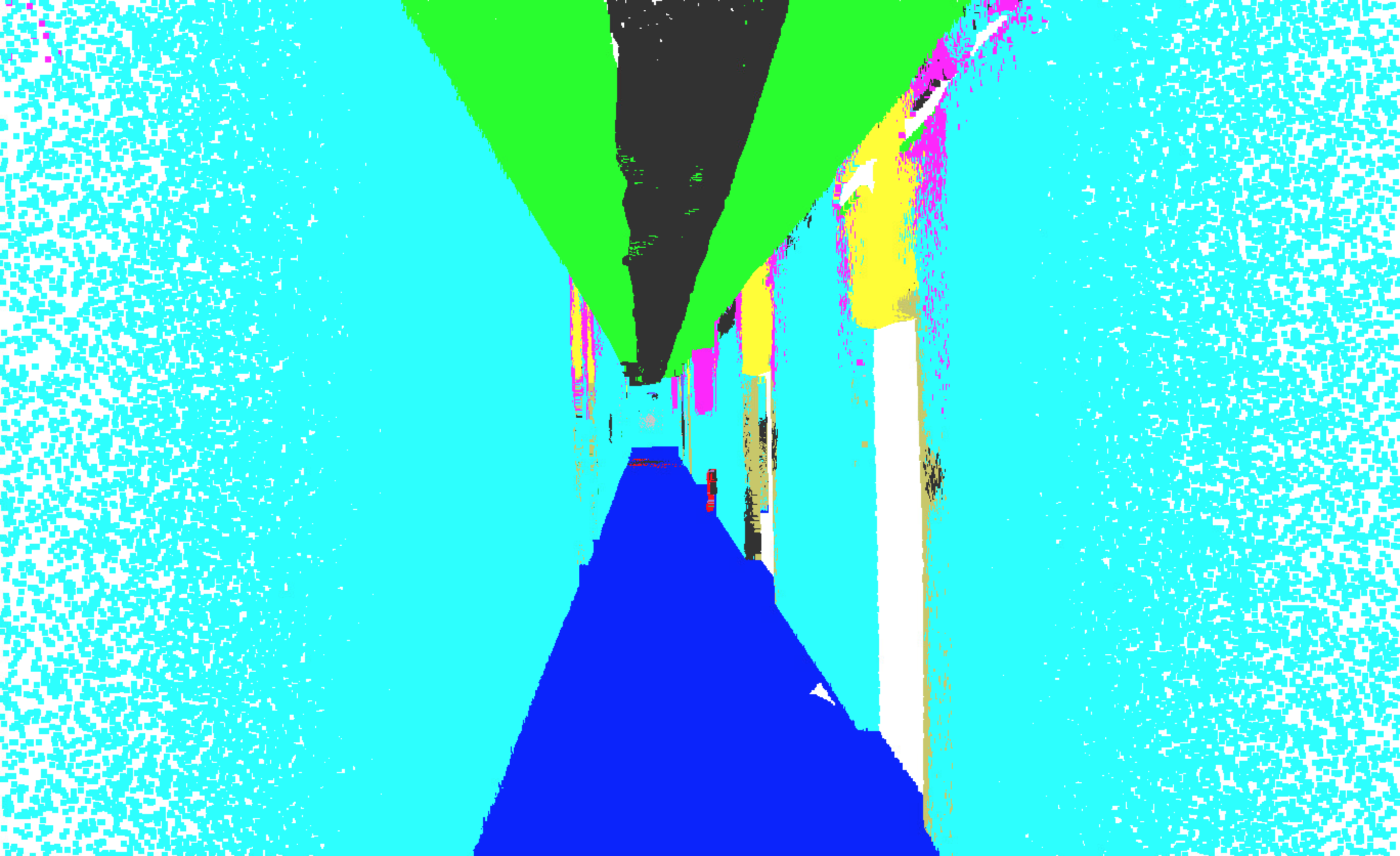}\hfill     
    \includegraphics[width=0.24\textwidth]{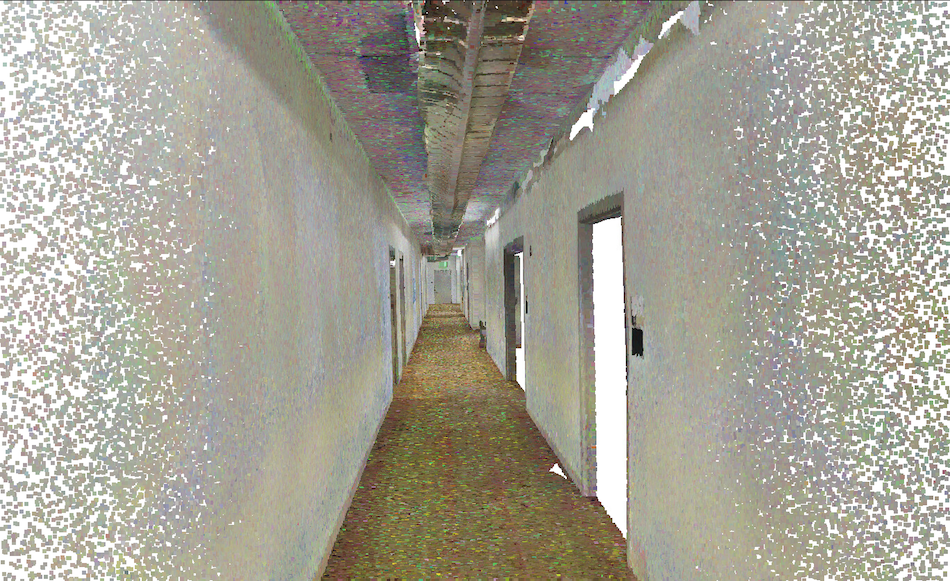}\hfill 
    \includegraphics[width=0.24\textwidth]{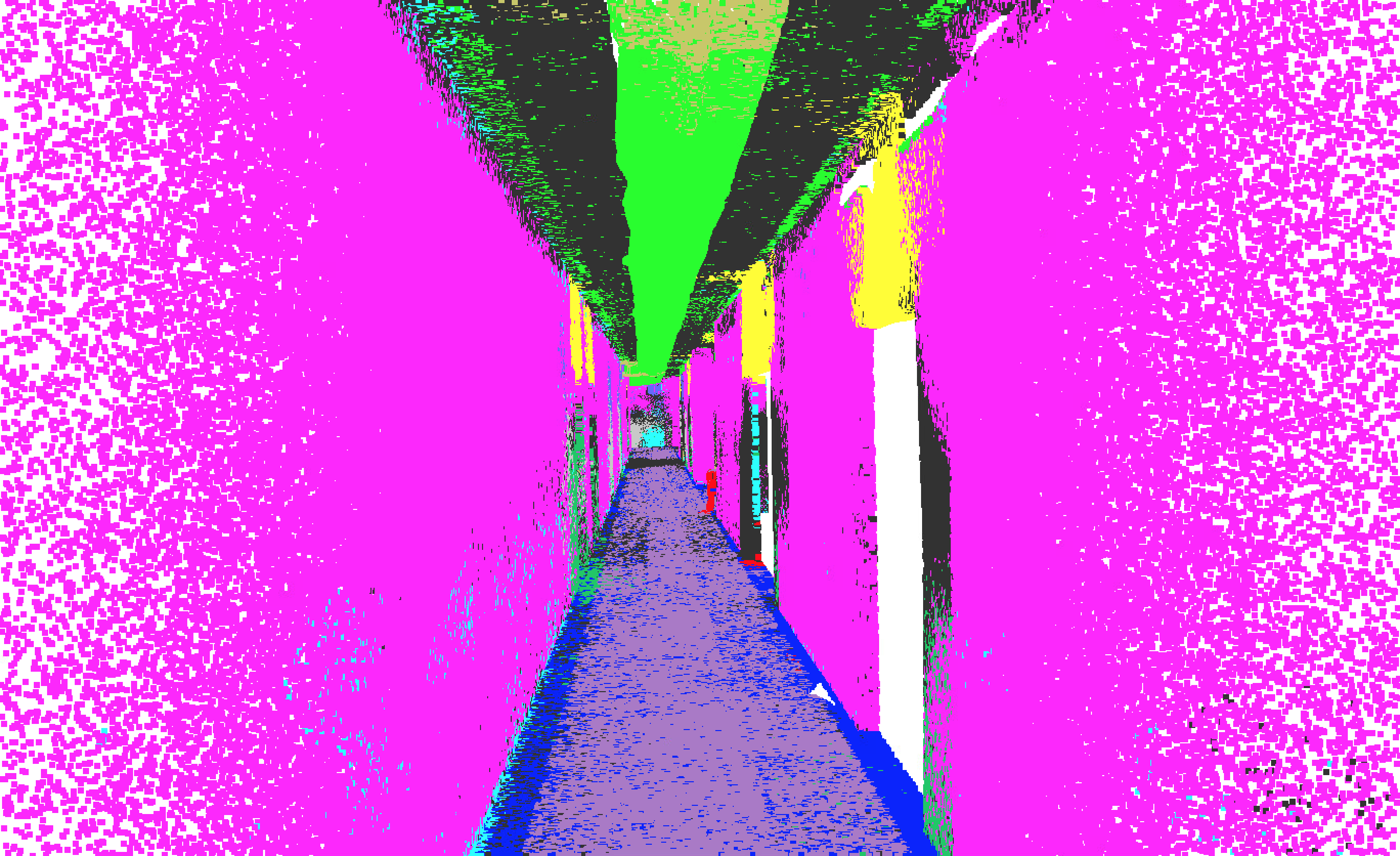}\hfill
    
    \includegraphics[width=0.24\textwidth]{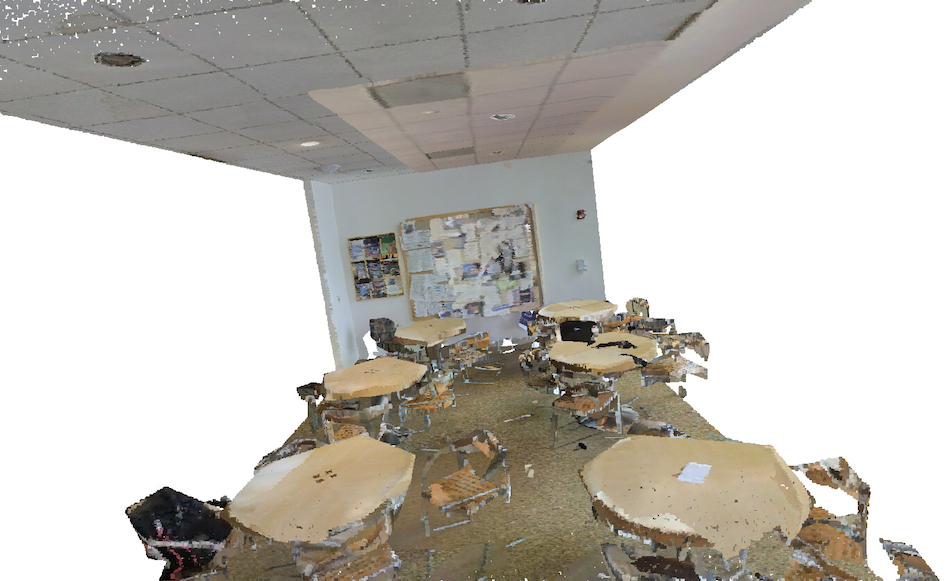}\hfill
    \includegraphics[width=0.24\textwidth]{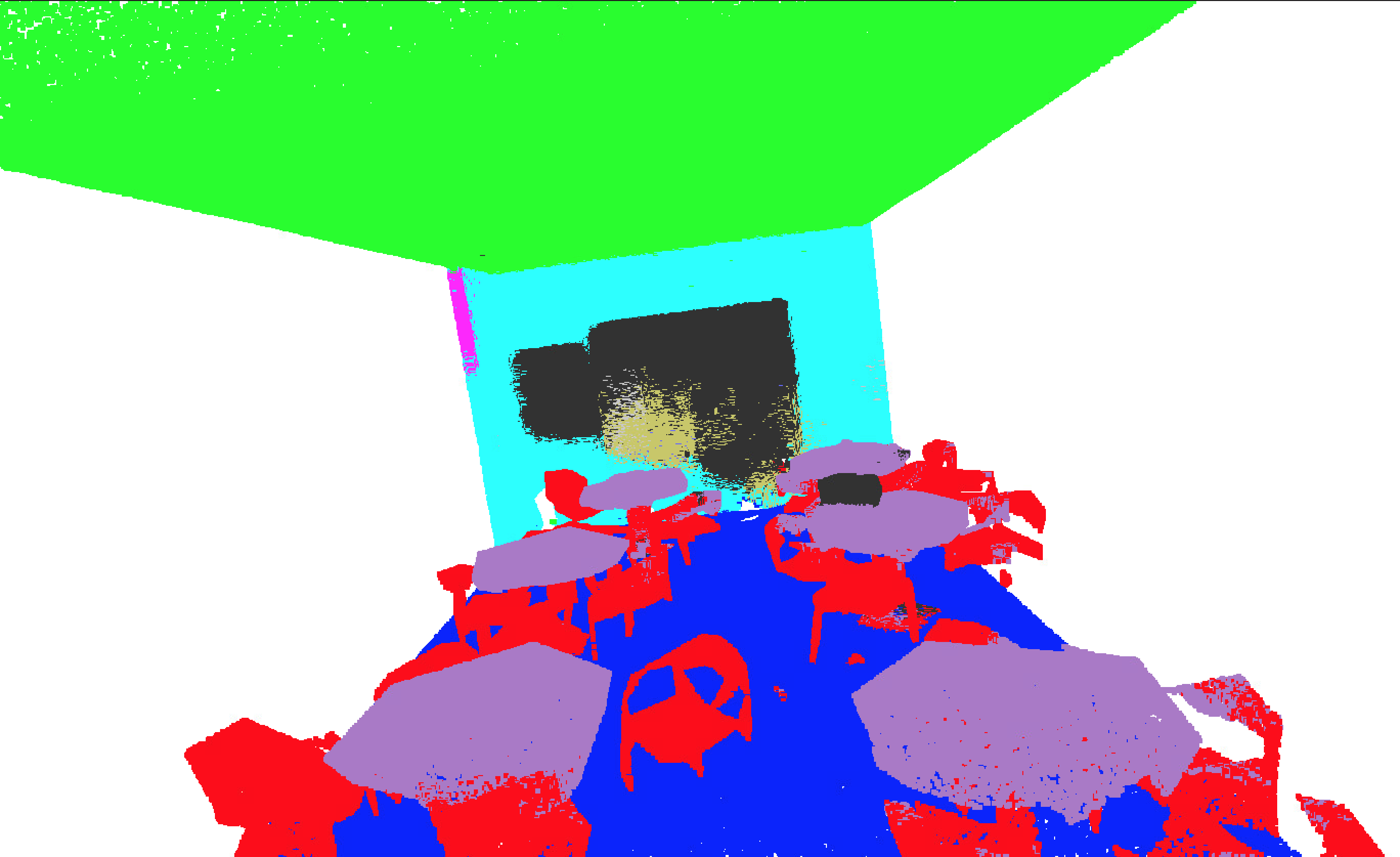}\hfill      
    \includegraphics[width=0.24\textwidth]{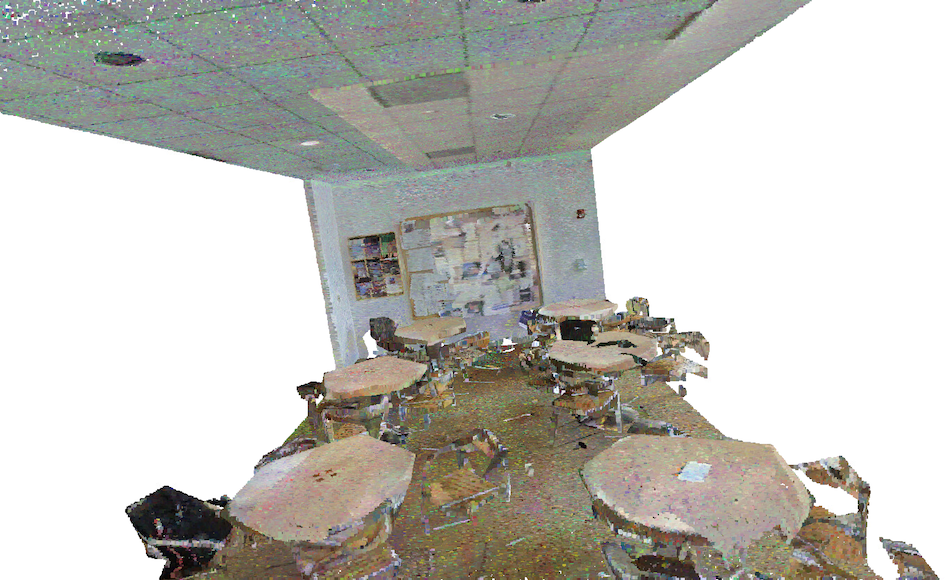}\hfill 
    \includegraphics[width=0.24\textwidth]{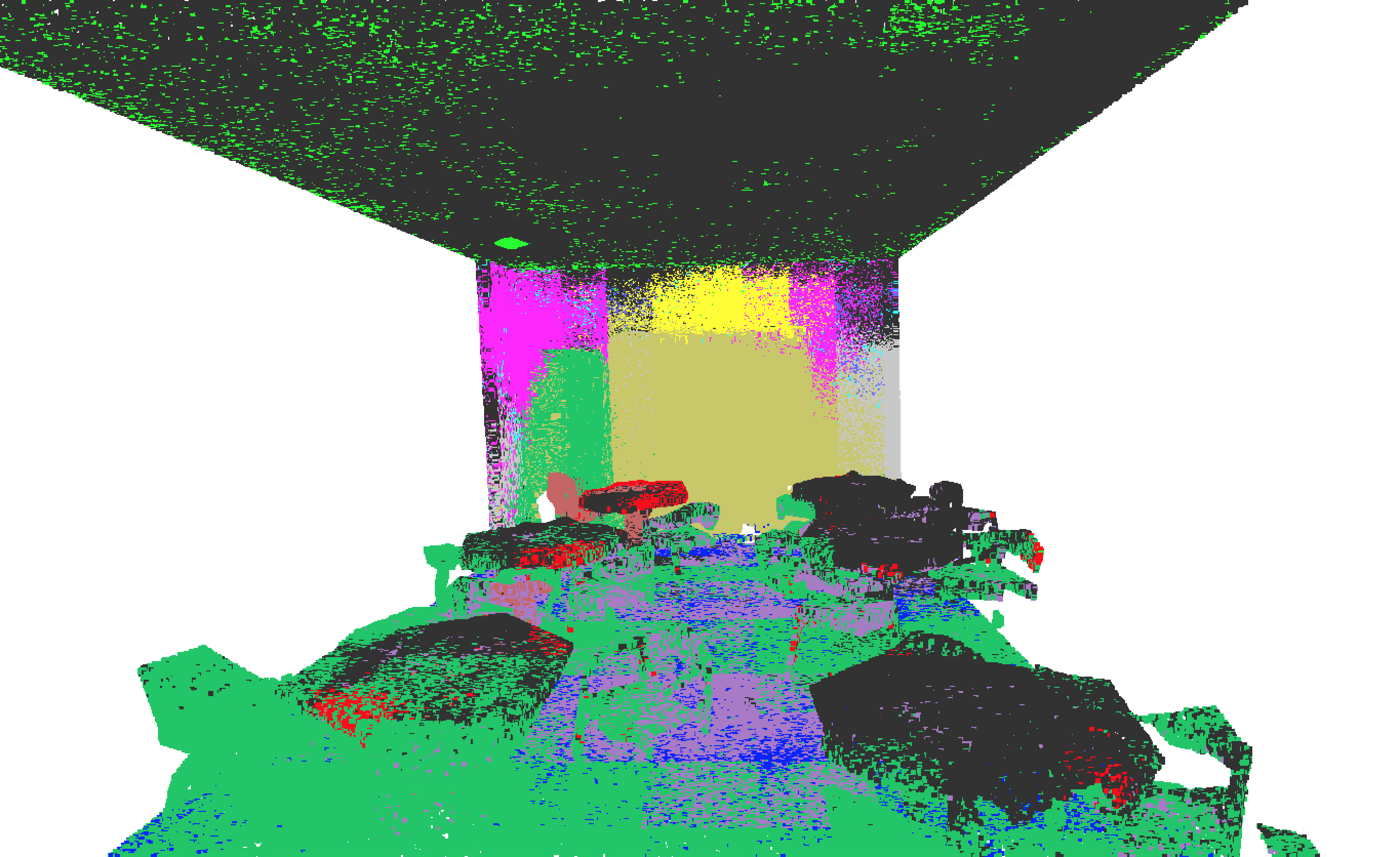}\hfill
    \caption{The Performance degradation attack with Conference room 1 (first row), Hallway 2 (second row), Lobby 1 (third row)  of Area 5 in S3DIS. The first to fourth columns show the original scene, the original segmentation results, perturbed scene and perturbed segmentation results.}
    \label{Fig:non_target_visualization_examples}
\end{figure*}

\begin{figure}[h]
    \includegraphics[width=0.49\columnwidth]{pics/raw.png}\hfill
    \includegraphics[width=0.49\columnwidth]{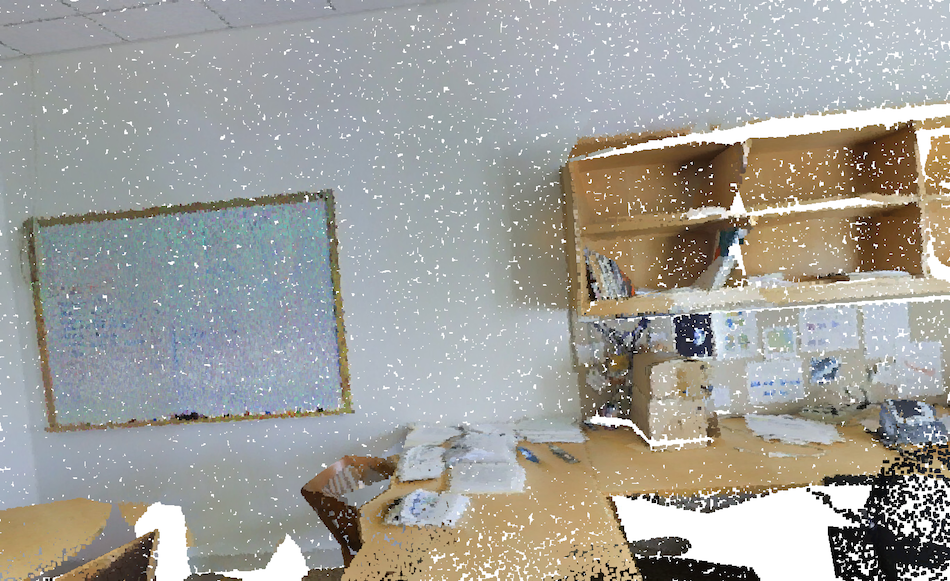}\hfill 
    \includegraphics[width=0.49\columnwidth]{pics/pred.png}\hfill
    \includegraphics[width=0.49\columnwidth]{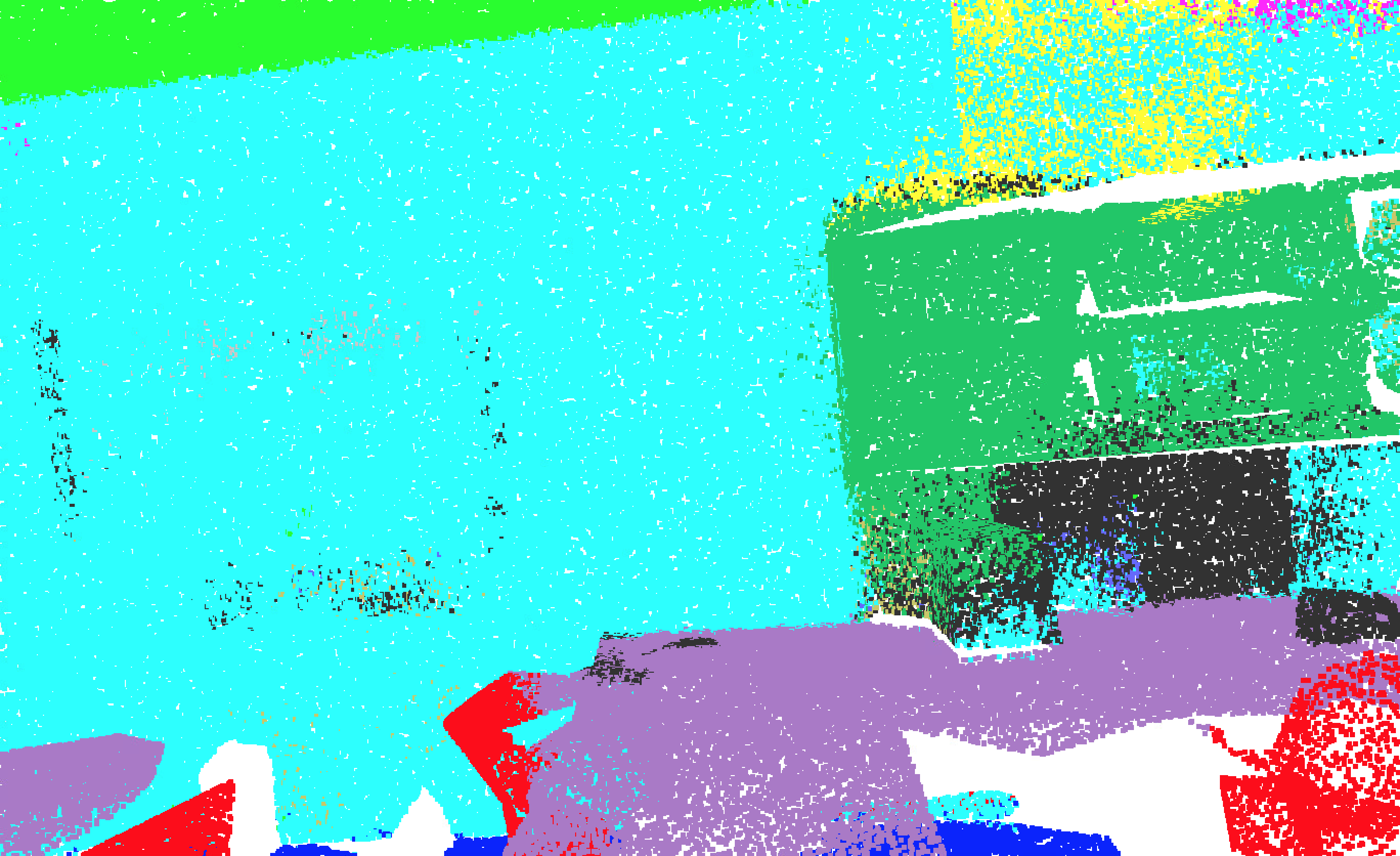}\hfill
    \caption{The object hiding attack with Office 33 of Area 5 in S3DIS. The upper left, upper right, lower left, and lower right show the original scene, the perturbed scene, and the segmentation results of the original and perturbed scene.}
    \label{Fig:target_example_s3dis_2}
\end{figure}
\ignore{
\begin{figure}[h]
    \includegraphics[width=0.49\columnwidth]{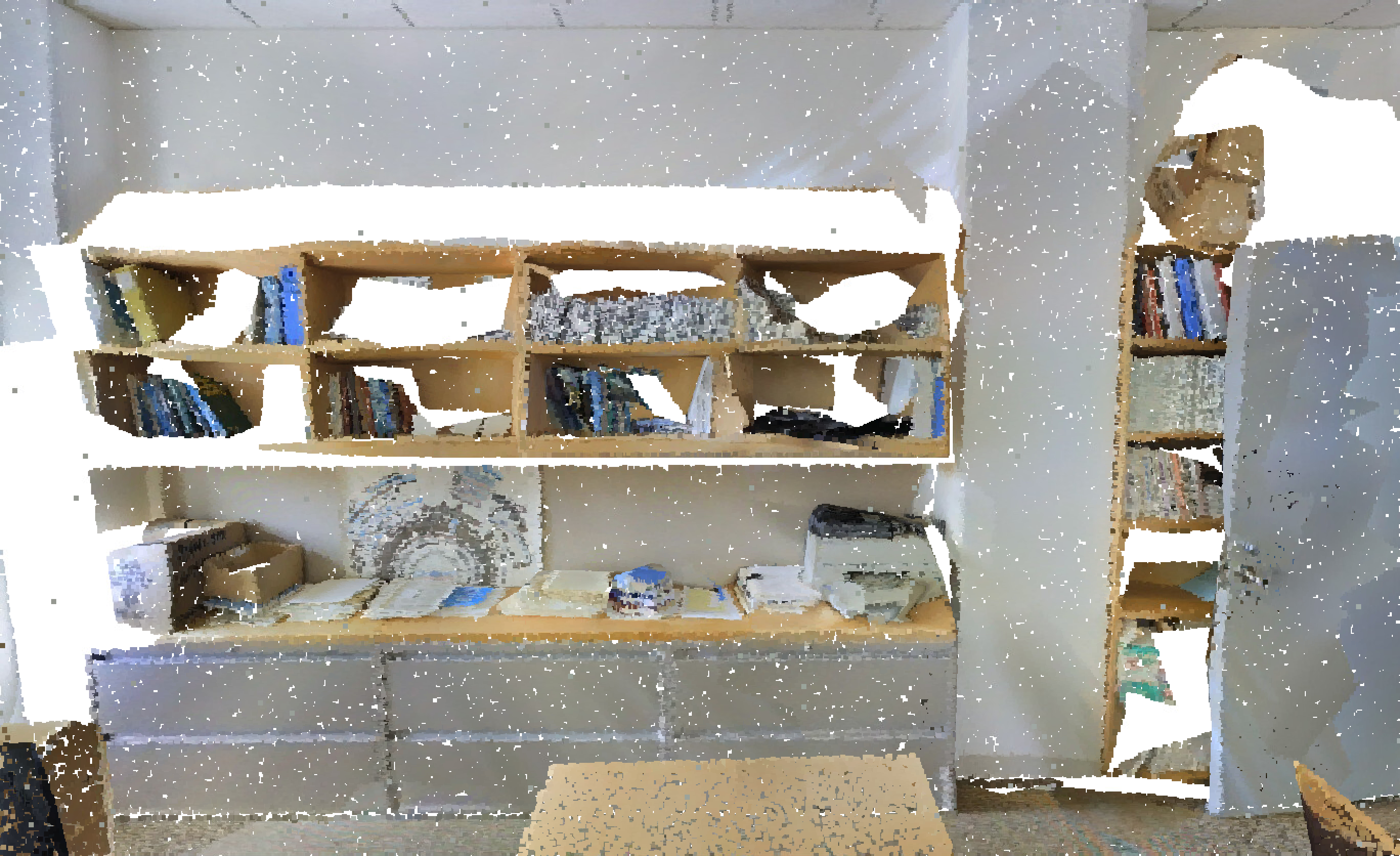}\hfill
    \includegraphics[width=0.49\columnwidth]{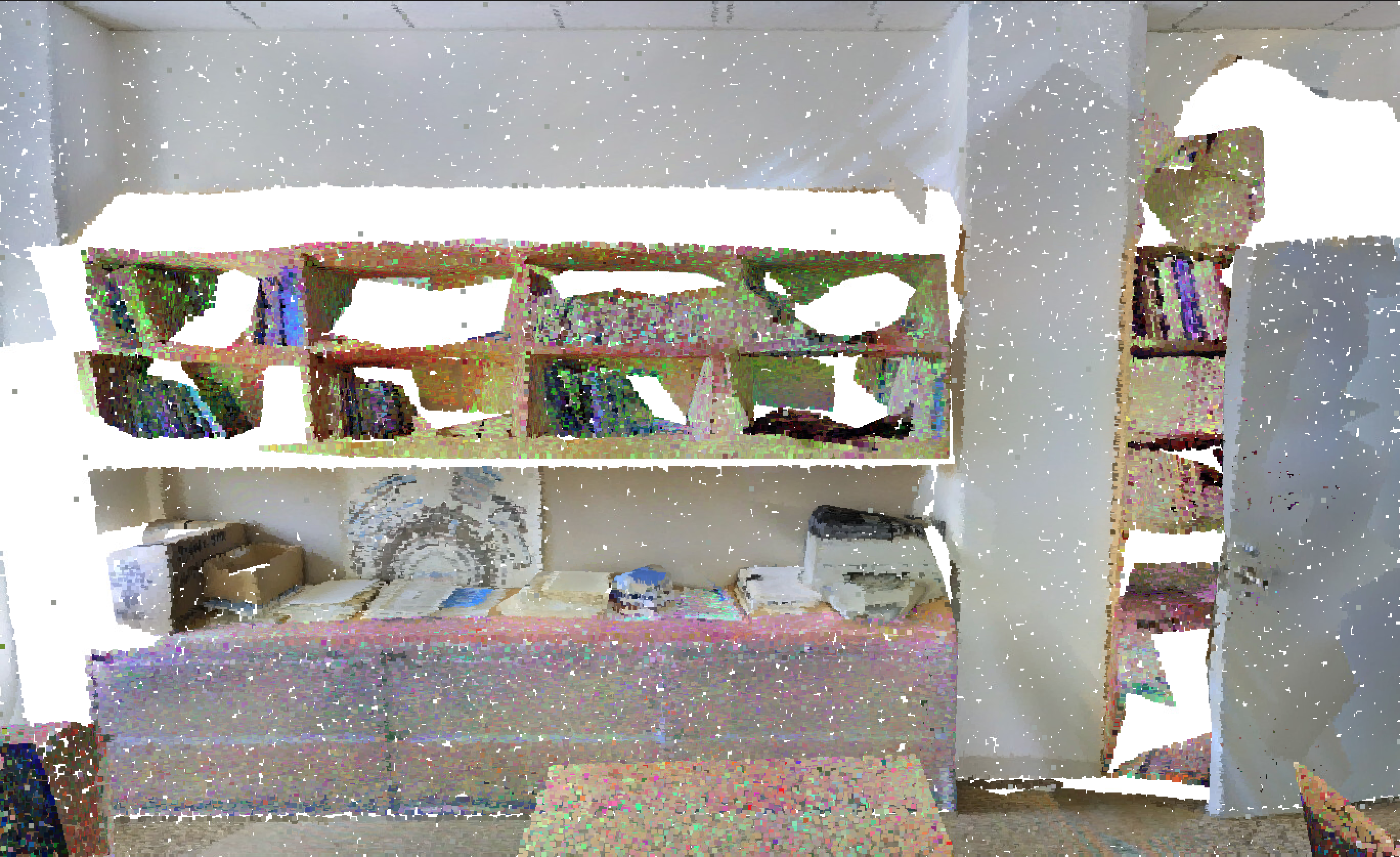} \hfill
    \includegraphics[width=0.49\columnwidth]{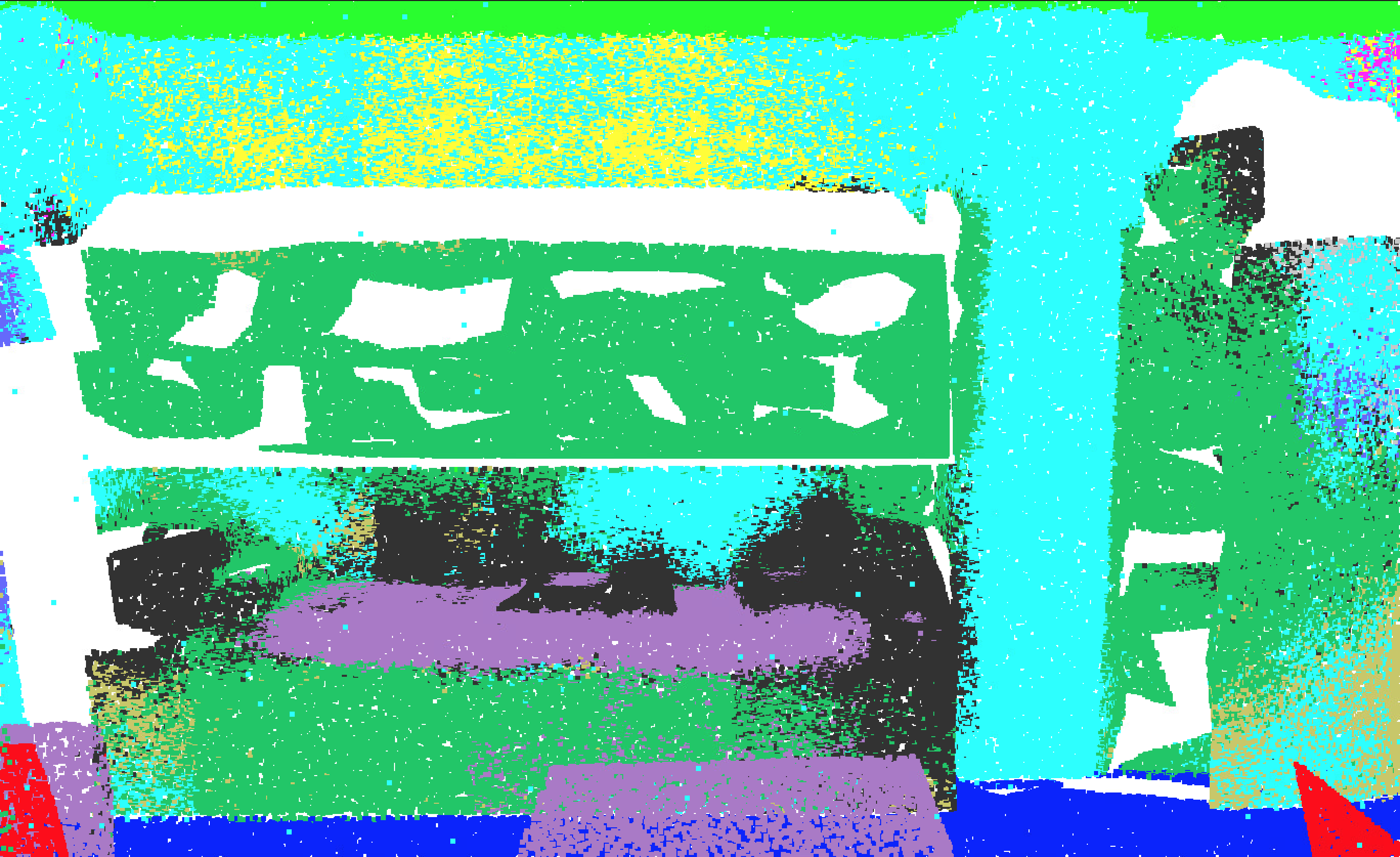}\hfill
    \includegraphics[width=0.49\columnwidth]{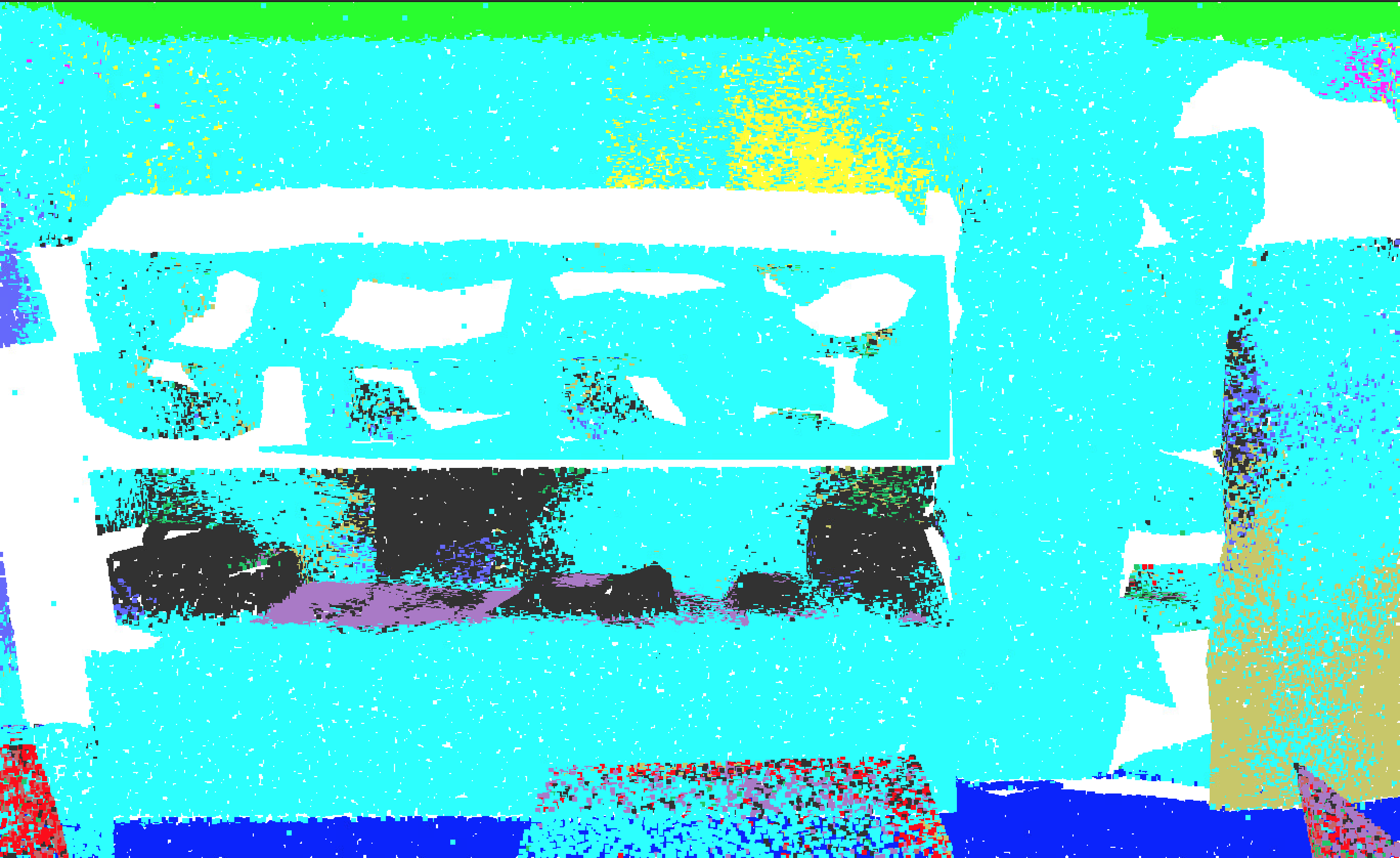}\hfill
    \caption{The object hiding attack against multiple source classes in Office 33 of Area 5 in S3DIS. Table, chair and bookcase are all misclassified as wall after the attack.}\label{Fig:target_example_s3dis_3}
\end{figure}
}

In this subsection, we visualize the adversarial examples generated under color-based norm-unbounded attack. For each sample, we show the original and perturbed scenes and their segmentation results. 

First, we show the scenes in S3IDS under performance degradation attack and PointNet++ is the target model. We choose different types of scenes like the conference room, hallway, and lobby. From Figure~\ref{Fig:non_target_visualization_examples}, we can see the small perturbation generated by our attack leads to prominent changes in the segmentation results. 

Next, we show an example of the object hiding attack in Figure~\ref{Fig:target_example_s3dis_2}. We set PointNet++ as the target model, and board as the source class. Since most of its points are classified as wall after the attack, our attack could make the board nearly ``disappear'' from the view of the segmentation model.

\ignore{
When multiple source classes are perturbed, our attack is demonstrated effective as well. 
We consider table (label=7), chair (label=8), and bookcase (label=10) as the source classes to be changed, and PointNet++ as the segmentation model. The result shows they are all mis-classified as wall by the segmentation model. 
}

\begin{figure}[h]
     \includegraphics[width=0.45\columnwidth]{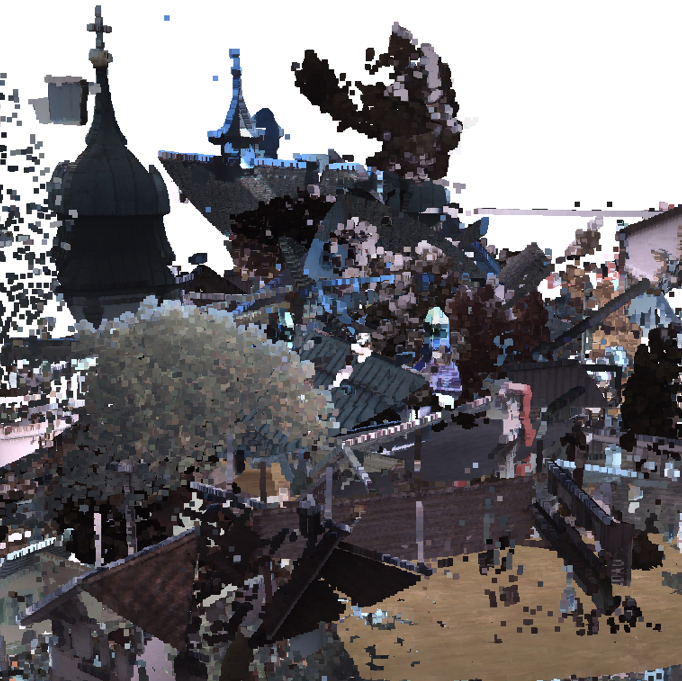}\hfill
     \includegraphics[width=0.45\columnwidth]{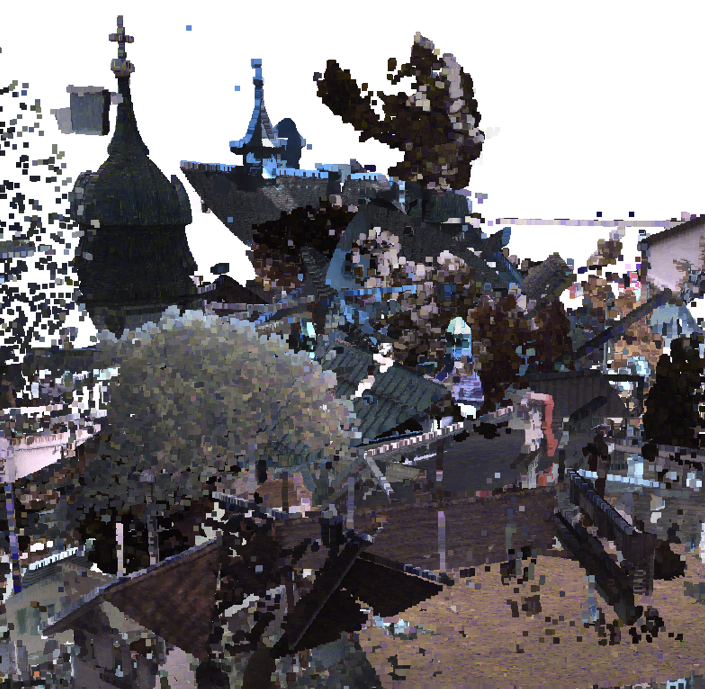}\hfill
     \includegraphics[width=0.45\columnwidth]{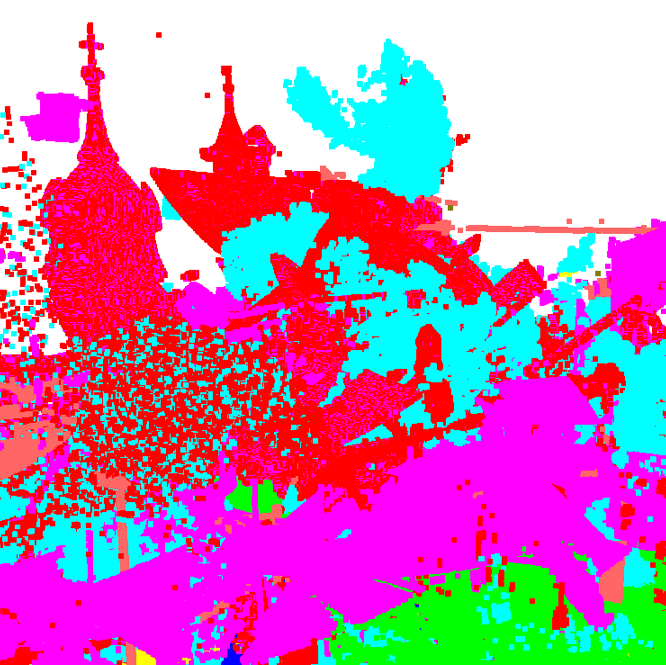}\hfill
     \includegraphics[width=0.45\columnwidth]{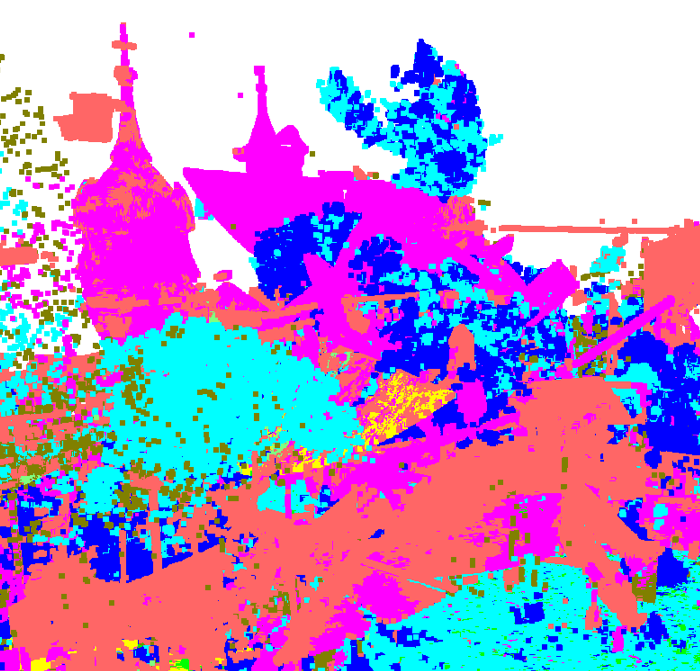}\hfill
    \caption{An example of performance degradation attack against Semantic3D.  }\label{Fig:non_target_example_semantic3d}
\end{figure}



Finally, we show an example of an outdoor scene under Semantic3D in 
Figure~\ref{Fig:non_target_example_semantic3d}, under performance degradation attack, with RandLA-Net as the targeted model. The visualized result also suggests seemingly small perturbations can drastically change the segmentation results.

\section{Discussion}
\label{sec:discussion}



\revised{
\zztitle{Sub-sampling}
Sub-sampling is a key technique in dealing with large-scale point cloud data, as described in Section~\ref{subsec:dl}. Defenses can also leverage sub-sampling, as shown in Section~\ref{subsec:defense}. We found when the sub-sampling is done by the PCSS models on the point cloud, such as farthest point sampling from PointNet++, $k$-neighbor Sampling from ResGCN and random sampling from RandLA-Net, our attacks are still effective.
When the sub-sampling is done before PCSS, e.g., storing a fraction of video frames from camera~\cite{tomei2019sensor}, our attacks could be impacted if the sampling procedure is unknown to the attacker, as the adversarial input cannot be directly constructed from the perturbed point cloud.
When sub-sampling is done as a defense, different sampling methods have different effectiveness (e.g., SOR is more effective against norm-unbounded attacks). 
\ignore{
(1) When sampling is done by the sensors ~\cite{}
data collection stage: sensors sample the points from the raw scenes and objects to generate point clouds. Data collection sampling does not impact the adversarial attacks since they target on the point clouds. 
(2) model sampling stage: different PCSS models use different sampling strategies to improve the efficiency.  are evaluated in the experiment. Our results shows that the model sampling strategies cannot improve the model robustness. 
(3) Defense sampling: SOR and SRS are two defense methods to evaluate our adversarial attack in Section~\ref{sec:evaluation}. We find that SOR can mitigate \CW\ attack to some degree while SRS does not work on both \CW\ and \PGD\ attacks.}
}

\zztitle{Other models}
\revised{
We select three representative PCSS models to attack. Section~\ref{subsec:dl} overviews the other types of models. We expect our attacks to be applicable to the models which generate gradients. One example is Point Cloud Transformer (PCT), which captures the context of a point with the Transformer architecture (e.g., through the positional encoder and self-attention)~\cite{guo2021pct}. PCT still computes gradients and recent works showed that Vision Transformer (ViT) in the 2D image domain is vulnerable under perturbation~\cite{wei2022towards, aldahdooh2021reveal}.
}



\zztitle{Limitations}
(1) Currently we evaluate the attacks on two datasets. 
Admittedly, we could extend the study scope by including more datasets.
(2) For RandLA-Net, we did not implement the coordinate-based attack as its point sampling mechanism makes it more difficult to locate the points for attack. 
(3) The focus of this study is to examine the robustness of different PCSS models and settings. Unlike previous papers~\cite{zhu2021adversarial, cao2021invisible, zhao2019seeing}, we did not convert the perturbation on the point cloud into the changes in the physical world, e.g., using irregular objects or stickers.
\revised{
(4) Our attacks target one point cloud at a time. In the real-world autonomous driving setting, a sequence of point clouds needs to be processed by a PCSS model, so the attacker should consider how to sufficiently attack multiple point clouds. Previous studies on the 2D image domain show that an attacker can add the same perturbation on multiple images, after assigning different weights to each image
~\cite{NEURIPS2021_85e5526a}. We expect a similar approach can be applied to 3D point clouds.
(5) Among the point cloud features, we only attack the color feature because it provides more information than the others (e.g., the intensity feature of Semantic3D). The distance function described in Section~\ref{subsec:components}  might need to be changed for other features.
}

\ignore{
To attack the autonomous system successfully, the point cloud sequence should be perturbed with the same noise. The min-max attack on vision has validated that  with the weights optimization depending on the robustness of each image. 
For the point cloud perturbation as mentioned in Section~\ref{sec:problem}, our attack methods can straightforwardly consider several point clouds by adding the losses together and assigning weights based on the point clouds' vulnerabilities. 
}




\section{Conclusion}
\label{sec:conclusion}

In this work, we present the first comparative study of adversarial attacks on 3D point cloud semantic segmentation (PCSS). 
We systematically formulate the attacker's objectives under the object hiding attack and the performance degradation attack, and develop two attack methods based on norm-bounded attack and norm-unbounded attack. In addition to the point coordinates that are exploited by all existing adversarial attacks, we consider point features to be perturbed. We examine these attack combinations on an indoor dataset S3IDS and an outdoor dataset Semantic3D dataset, to examine the impact of each attack option. 
Overall, we found all examined PCSS models are vulnerable under adversarial perturbation, in particular to norm-unbounded attack that is applied to the color features. 
We hope with this study, more efforts can be made to improve the robustness of PCSS models.

\ignore{To highlight our evaluation result, we found feature-based attack, especially color-based attack, is more effective than coordinate-based attack. Norm-unbounded attack is more effective, driving down the accuracy and aIoU by a big margin, and the result is consistent across scenes and the targeted models. 
For the targeted attack, we observe the prominent differences between objects, and how to make the targeted attack effective across different classes warrant future research. All targeted models, including PointNet++, ResGCN-28 and RandLA-Net, are vulnerable, and the adversarial examples are transferrable. 
}


 \section*{Acknowledgment}
We thank the valuable comments from the anonymous reviewers and our shepherd. The authors also thank Lijie Huang from the UCInspire program for the help.
The authors from the United States are supported under NSF 2039634 and 2124039. The author from China is supported under National Key R\&D Program of China (Grant No.2022YFB3102901) and the Natural Science Foundation of Shanghai (No. 23ZR1407100).


\bibliographystyle{plain}

\bibliography{main}

\clearpage

\newpage

\ignore{

\section{The summary of the previous review}
Our paper was submitted to AAAI'22. We made major changes to the last version in light of reviewers' comments. Below summarize the changes. The original review is attached after the summary.

\begin{itemize}
    \item Comparing to the previous submission, we change the paper theme from designing a color-only attack to a comparative evaluation of the semantic segmentation models (PointNet++, ResGCN-28, and RandLA-Net).
    \item We added the norm-bounded attack in addition to the norm-unbounded attack  from the last version, and we define their objectives in Section~\ref{sec:problem} in a more consistent way.
    \item We summarize the related works on both the deep learning on point clouds and adversarial attack directions with newly published works.
    \item We describe the details and compare the differences between norm-bound and norm-unbound attacks with the support of algorithms and figures.
    \item We evaluate two attacks on the S3DIS dataset to compare the color and coordinate features on S3DIS.
    \item We highlight the research questions (RQ) and answer them accordingly.
\end{itemize}

}


\end{document}